\documentclass{article}

\usepackage[preprint]{neurips_2024}

\usepackage{amsmath}
\usepackage{amssymb}
\usepackage{mathtools}
\usepackage{amsthm}
\usepackage{multirow}
\usepackage{algorithm}
\usepackage{algpseudocode}
\usepackage{todonotes}
\usepackage{subcaption}
\usepackage{pifont}
\newcommand{\cmark}{\ding{51}}
\newcommand{\xmark}{\ding{55}}

\usepackage{xcolor}

\usepackage[utf8]{inputenc} 
\usepackage[T1]{fontenc}    
\usepackage{hyperref}       
\usepackage{url}            
\usepackage{booktabs}       
\usepackage{amsfonts}       
\usepackage{nicefrac}       
\usepackage{microtype}      
\usepackage{xcolor}         

\usepackage{adjustbox}

\title{Computational Tradeoffs in Image Synthesis: Diffusion, Masked-Token, and Next-Token Prediction}

\author{Maciej Kilian$^\pi$
$\quad$
Varun Jampani$^\pi$ 
$\,\,$
Luke Zettlemoyer$^{\mu}$ 
\\
$^\pi$StabilityAI $\qquad$ $\qquad$ $\qquad$ $^\mu$University of Washington \\
}

\begin{document}

\maketitle

\newcommand{\aereconstructiontable}{
\begin{table}[t]
\begin{minipage}{.5\textwidth}
  \centering
  \begin{tabular}{lccc}
    \toprule
    Regularizer & Latent space capacity & rFID ($\downarrow$) \\
    \midrule
    KL & 16 channels & 1.060 \\
    KL & 8 channels & 1.560 \\
    KL$_{es}$ & 8 channels & 2.856 \\
    KL & 4 channels & 2.410 \\
    LFQ & 16384 vocabulary & 2.784 \\
    \bottomrule \\
  \end{tabular}
  \caption{\textbf{Autoencoders.} Reconstruction metrics for differently regularized and trained autoencoders (downsampling $f=8$). "es" is early stopping to match the LFQ autoencoder. \vspace{-1em}}
  \label{tab:aereconstructiontable}
  \end{minipage}
  \hspace{0.2cm}
  \begin{minipage}{.5\textwidth}
  \centering
    \small
    \begin{tabular}{l c c c c c}
    \toprule
    Model size & Layers $N$ & Hidden size $d$ &  Heads \\
    \midrule 
    S  &   12   &      768    &   12  \\
    M  &    24   &      1024    &   16  \\
    L &    24  &       1536     &   16  \\
    XL & 32 & 2304 & 32 \\
    \bottomrule \\
    \end{tabular}
    \caption{\textbf{Transformer configurations.} Base transformer hyperparameters for models we train. Common across all approaches}
    \label{tab:tfsizetable}
  \end{minipage}
  \vspace{-1em}
\end{table}
}

\newcommand{\modelflopstable}{
\begin{table}
\begin{minipage}{.5\textwidth}
\centering
\small
\scalebox{0.9}{
\begin{tabular}{l c c c}
\toprule
Model & n-parameters (\%) &  Forward TFLOPs \\
\midrule 
DiT-S  &   131.13 M (97.2\%)   &   0.2133  \\
DiT-M  &   459.19 M (98.7\%)   &   0.7234  \\
DiT-L  &   1031.67 M (98.8\%) &   1.5485  \\
DiT-XL & 3083.69 M (99.2\%) & 4.4901 \\
NT/MT-S  &   153.73 M (82.9\%)   &   0.2261  \\
NT/MT-M  &   494.56 M (92.9\%)   &   0.6631  \\
NT/MT-L  &   1072.14 M (95.1\%) &   1.3421  \\
NT/MT-XL & 3137.29 M (97.5\%) & 3.7166 \\
\bottomrule \\
\end{tabular}}
\caption{\textbf{Model forward pass costs.} Number of parameters and FLOPs used in each forward pass for all models we trained. DiT - Diffusion Transformer; NT - Next-token; MT - Masked-token}
\label{tab:modelflopstable}
\end{minipage}%
\hspace{0.2cm}
\begin{minipage}{.5\textwidth}
\small
\scalebox{0.9}{
\begin{tabular}{l c c c c c}
\toprule
Objective & Conditioning & FID &  CLIP \\
\midrule 
NT  &   adaLNzero   &      \textbf{83.052}    &   \textbf{0.2213}  \\
NT  &    in context   &      88.176    &   0.2041  \\
NT &    cross attention  &       92.852     &   0.2062  \\
\midrule
MT  &   adaLNzero   &      \textbf{97.021}    &   \textbf{0.2164}  \\
MT  &    in context   &      100.646    &   0.1925  \\
MT &    cross attention  &       103.221     &   0.1960  \\
\bottomrule \\
\end{tabular}}
\caption{\textbf{Conditioning method ablation.} Results for different objectives and conditioning methods. adaLNzero conditioning is used for the remainder of experiments. NT - Next-token; MT - Masked-token.}
\label{tab:condablation}
\end{minipage}
\vspace{-2em}
\end{table}
}

\newcommand{\emaablation}{
\begin{table}
\centering
\small
\scalebox{0.9}{
\begin{tabular}{l c c c c c}
\toprule
Objective & LR schedule & EMA & FID &  CLIP \\
\midrule 
Next-token & constant & \xmark &  81.976    &   0.2208  \\
Next-token & constant & \cmark &   79.571   &   0.2230  \\
Next-token & cosine & \xmark &    \textbf{75.715}    &   0.2256  \\
Next-token & cosine & \cmark &   76.404   &   \textbf{0.2257}  \\
\midrule
Diffusion & constant & \xmark &  74.087    &   0.2153  \\
Diffusion & constant & \cmark &   71.789   &   0.2166  \\
Diffusion & cosine & \xmark &    \textbf{69.284}    &   \textbf{0.2195}  \\
Diffusion & cosine & \cmark &   69.468   &   0.2192  \\
\bottomrule
\\
\end{tabular}}
\caption{\textbf{EMA and learning rate schedules.} EMA on model weights improves results under a constant learning rate schedule but does not exceed the gains from using a cosine decay schedule.}
\label{tab:emaablation}
\end{table}
}

\newcommand{\diffae}{
\begin{figure*}[t]
\centering
\begin{subfigure}[t]{0.49\textwidth}
    \centering
    \includegraphics[width=0.9\linewidth]{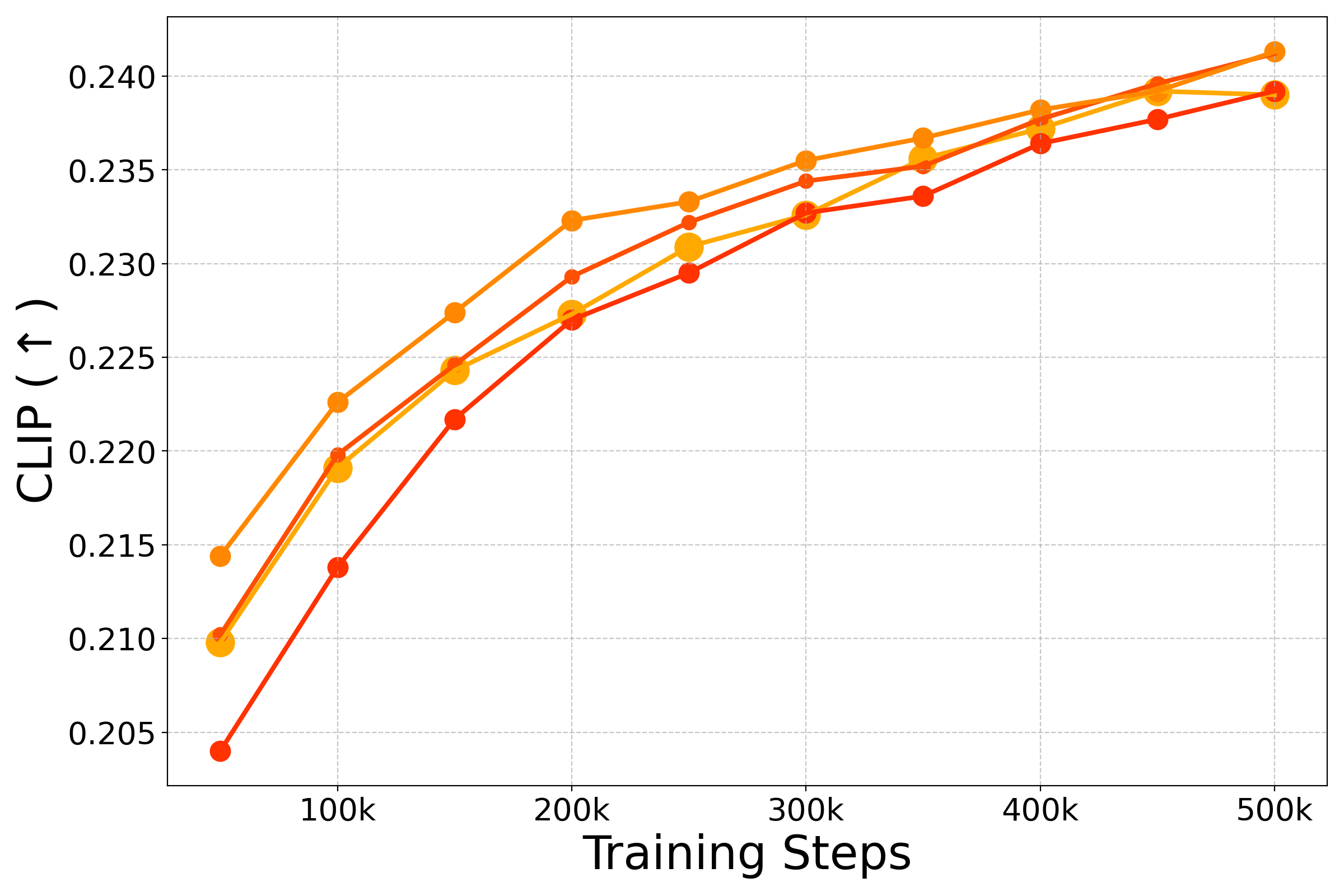}
    \label{fig:diffae:clip}
\end{subfigure}\hfill%
\begin{subfigure}[t]{0.49\textwidth}
    \centering
    \includegraphics[width=0.9\linewidth]{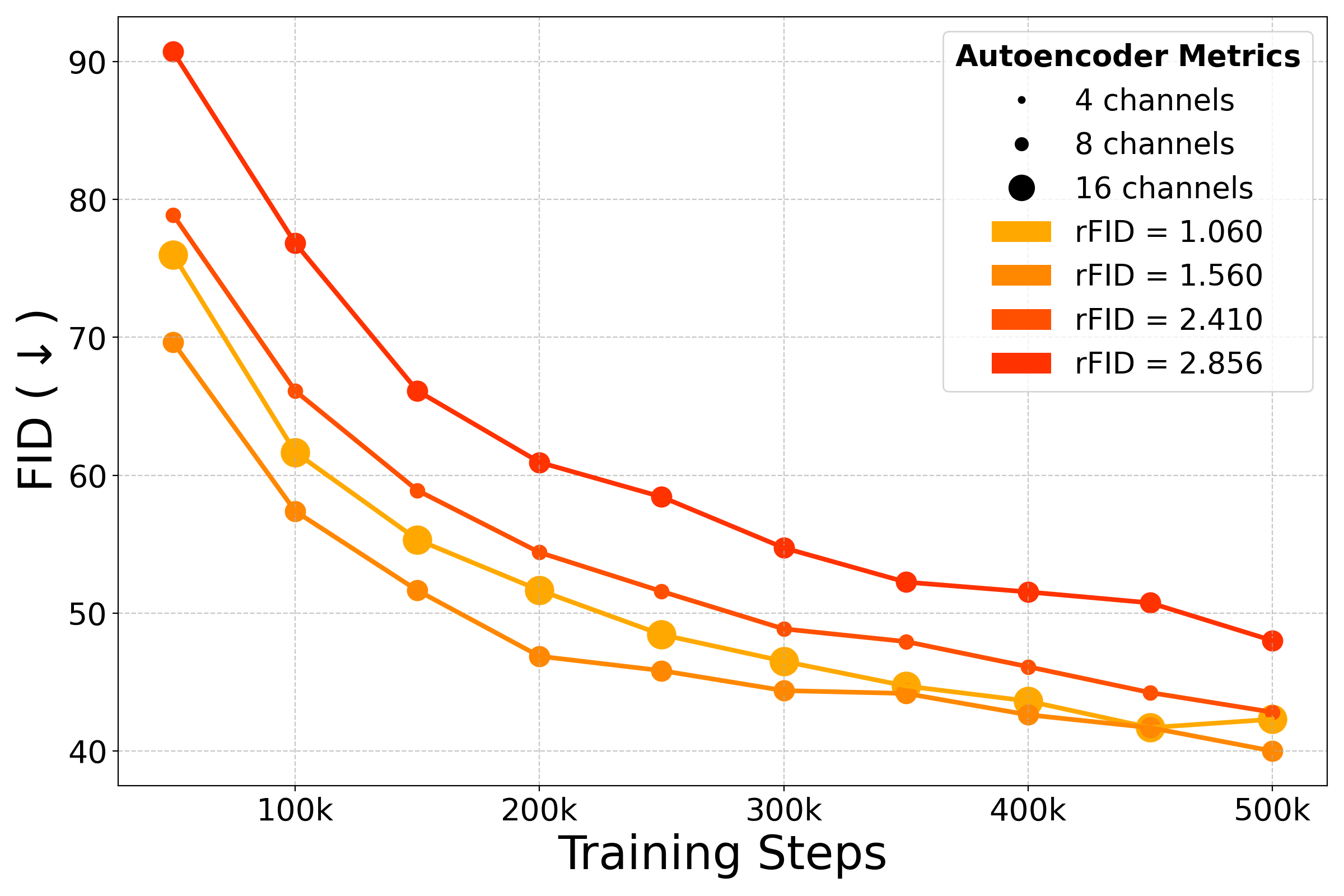}
    \label{fig:diffae:fid}
\end{subfigure}
\caption{\textbf{Impact of autoencoder quality on diffusion models.} We train L-size diffusion models on our set of continuous latent space autoencoders. The choice of autoencoder has more impact on FID than CLIP score.  Effectively using a larger latent space requires more compute and model capacity.}
\vspace{-1em}
\label{fig:diffae}
\end{figure*}
}

\newcommand{\mainres}{
\begin{figure*}[t]
\centering
\begin{subfigure}[t]{0.49\textwidth}
    \centering
    \includegraphics[width=\linewidth]{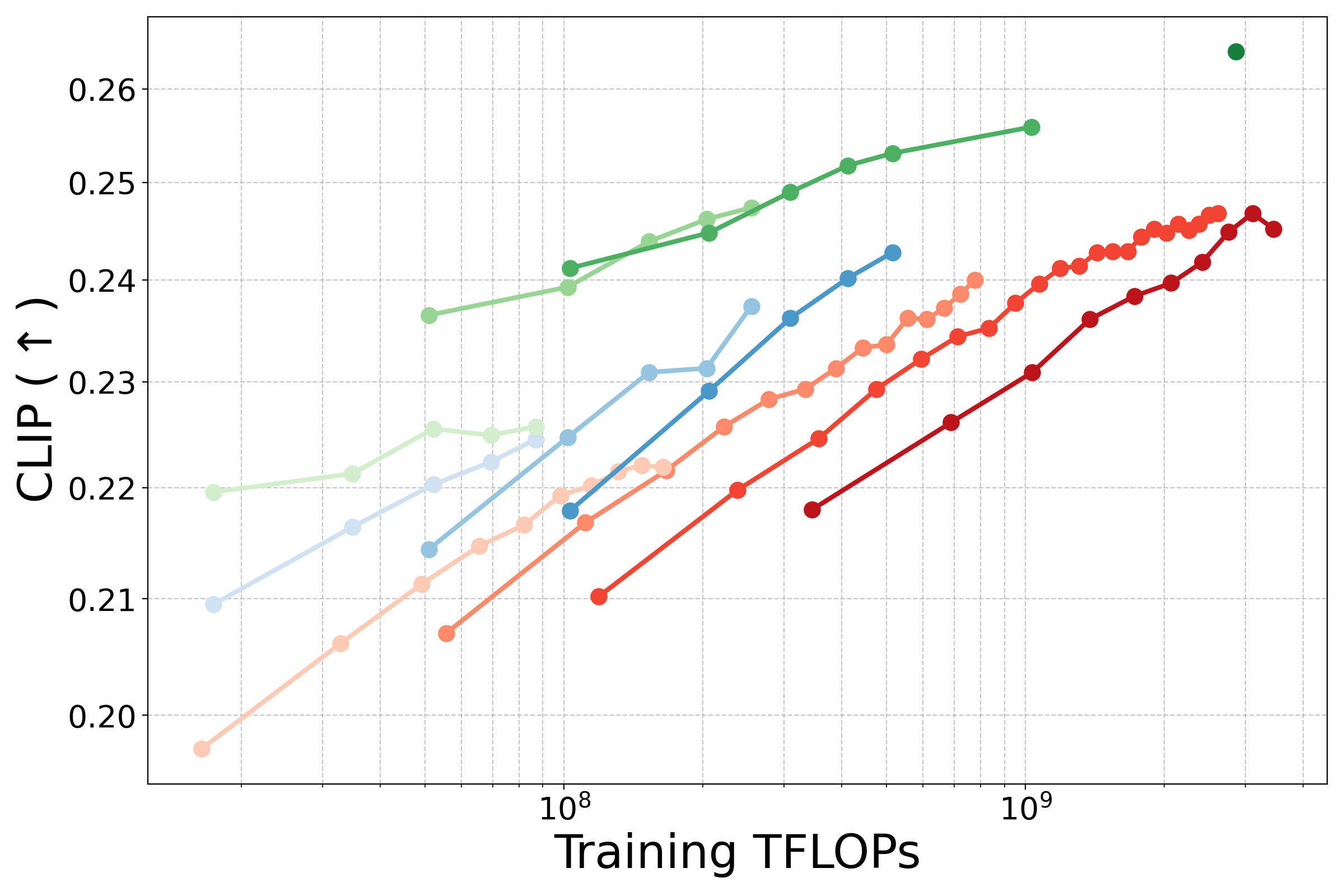}
    \label{fig:main:clip}
\end{subfigure}\hfill%
\begin{subfigure}[t]{0.49\textwidth}
    \centering
    \includegraphics[width=\linewidth]{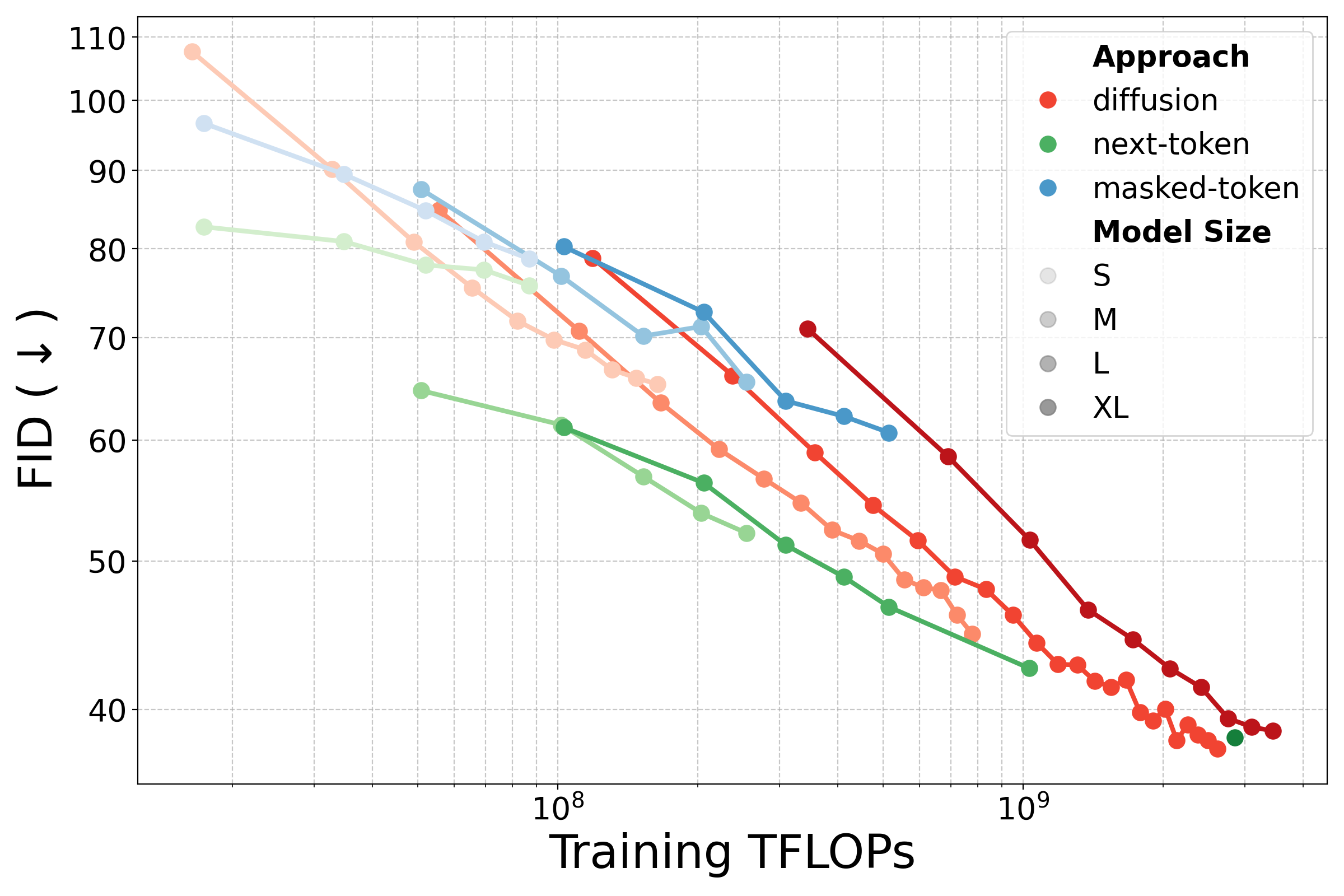}
    \label{fig:main:fid}
\end{subfigure}
\vspace{-1em}
\caption{\textbf{Training compute efficiency on perceptual metrics.} Performance on CLIP and FID scores for various models and dataset sizes across different image synthesis approaches. On FID, next-token prediction is initially the most compute-efficient but scaling trends suggest it is eventually matched by diffusion. Token-based methods significantly outperform diffusion in CLIP score. Both axes are in log scale.
\vspace{-1em}}
\label{fig:main}
\end{figure*}
}

\newcommand{\fintl}{
\begin{figure*}[t]
\centering
\includegraphics[width=0.9\linewidth]{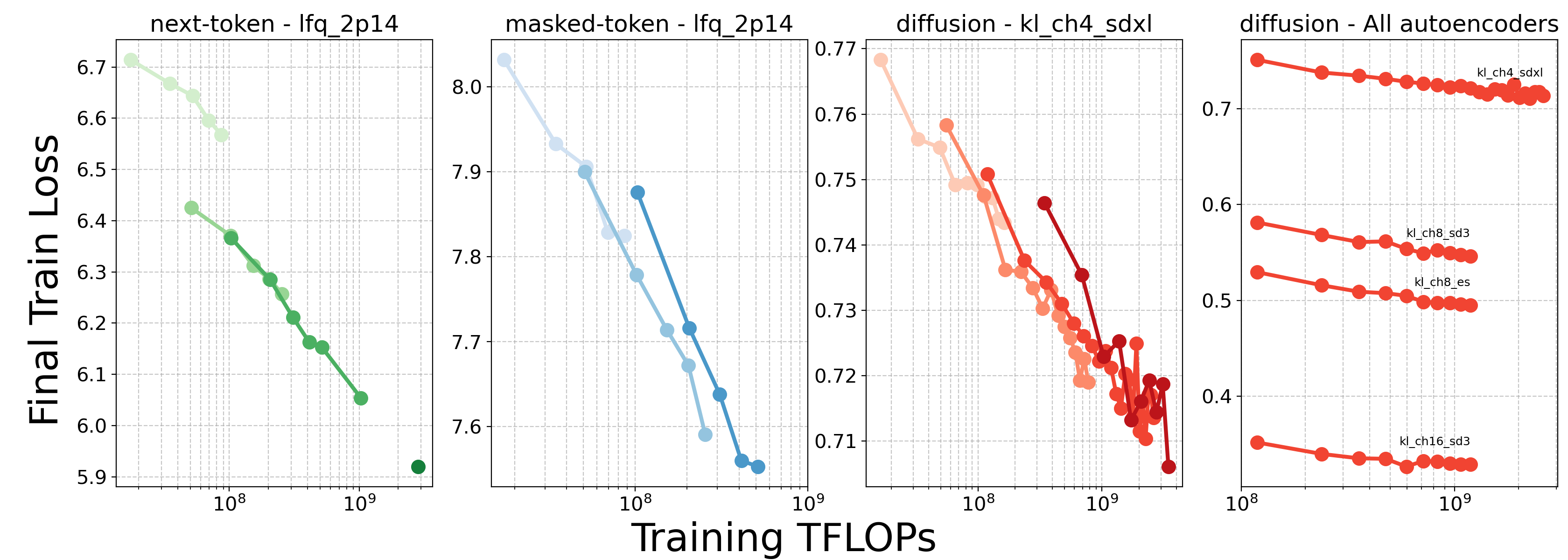}
\caption{\textbf{Training compute efficiency on final loss.} All objectives follow predictable scaling trends. Right plot shows the difference in loss scale between diffusion models trained on top of different autoencoders. FLOPs axis is in log scale.
\vspace{-1em}}
\label{fig:fintl}
\vspace{-1em}
\end{figure*}
}

\newcommand{\sdres}{
\begin{figure*}[t]
\centering
\begin{subfigure}[t]{0.49\textwidth}
    \centering
    \includegraphics[width=\linewidth]{img/sd3_clip.png}
    \label{fig:sdres:clip}
\end{subfigure}\hfill%
\begin{subfigure}[t]{0.49\textwidth}
    \centering
    \includegraphics[width=\linewidth]{img/sd3_fid.png}
    \label{fig:sdres:fid}
\end{subfigure}
\caption{\textbf{Skewing Autoencoding Conditions.} Beyond a specific compute budget, diffusion models, when equipped with a superior autoencoder exceed the FID compute efficiency of next-token prediction. However, in terms of CLIP scores, token-based methods continue to significantly outperform. 
\vspace{-1em}}
\label{fig:sdres}
\end{figure*}
}

\newcommand{\emainfluence}{
\begin{figure*}[t]
\centering
\begin{subfigure}[t]{0.5\textwidth}
    \centering
    \includegraphics[width=\linewidth]{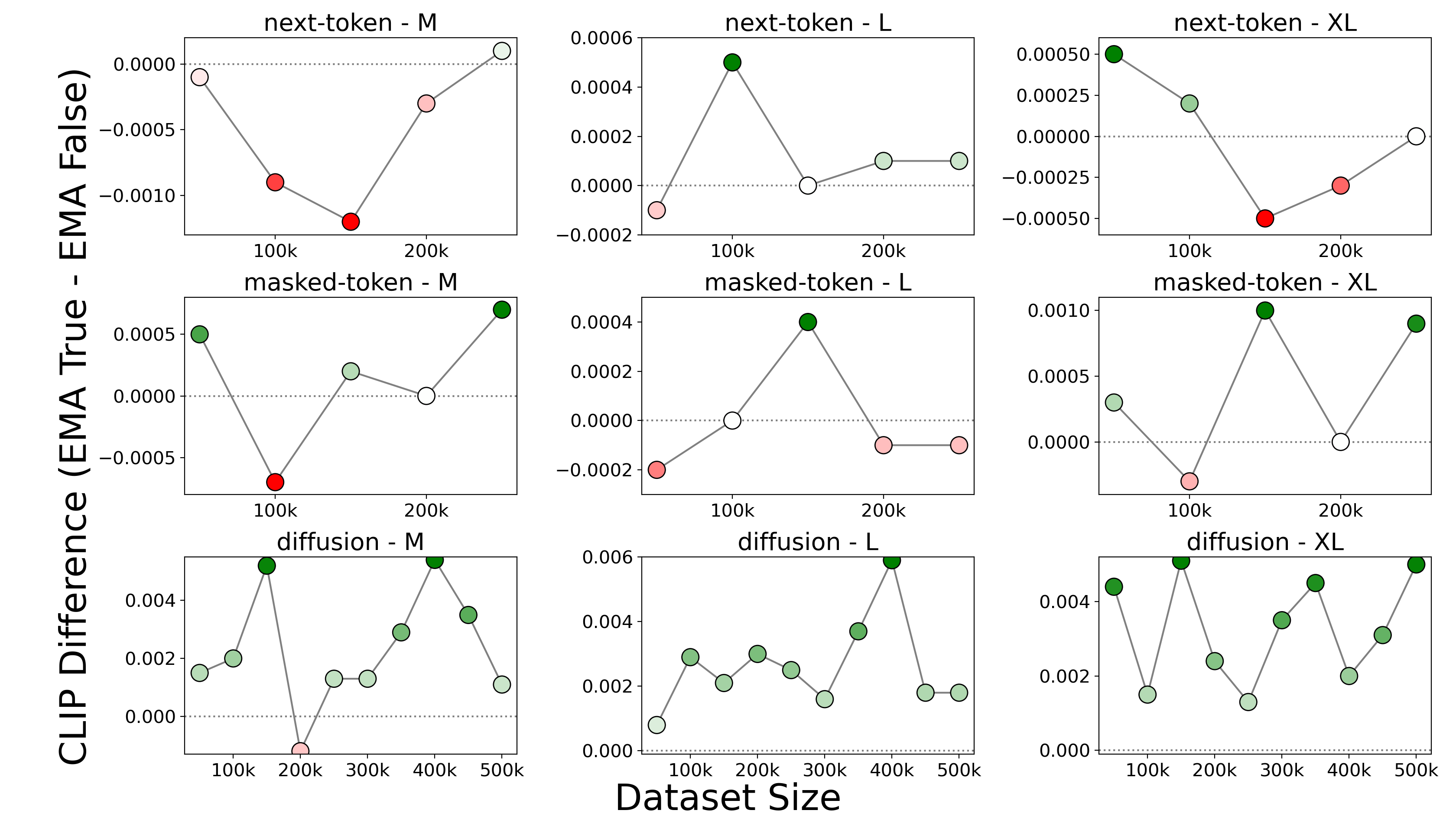}
    \label{fig:sd:clip}
\end{subfigure}\hfill%
\begin{subfigure}[t]{0.5\textwidth}
    \centering
    \includegraphics[width=\linewidth]{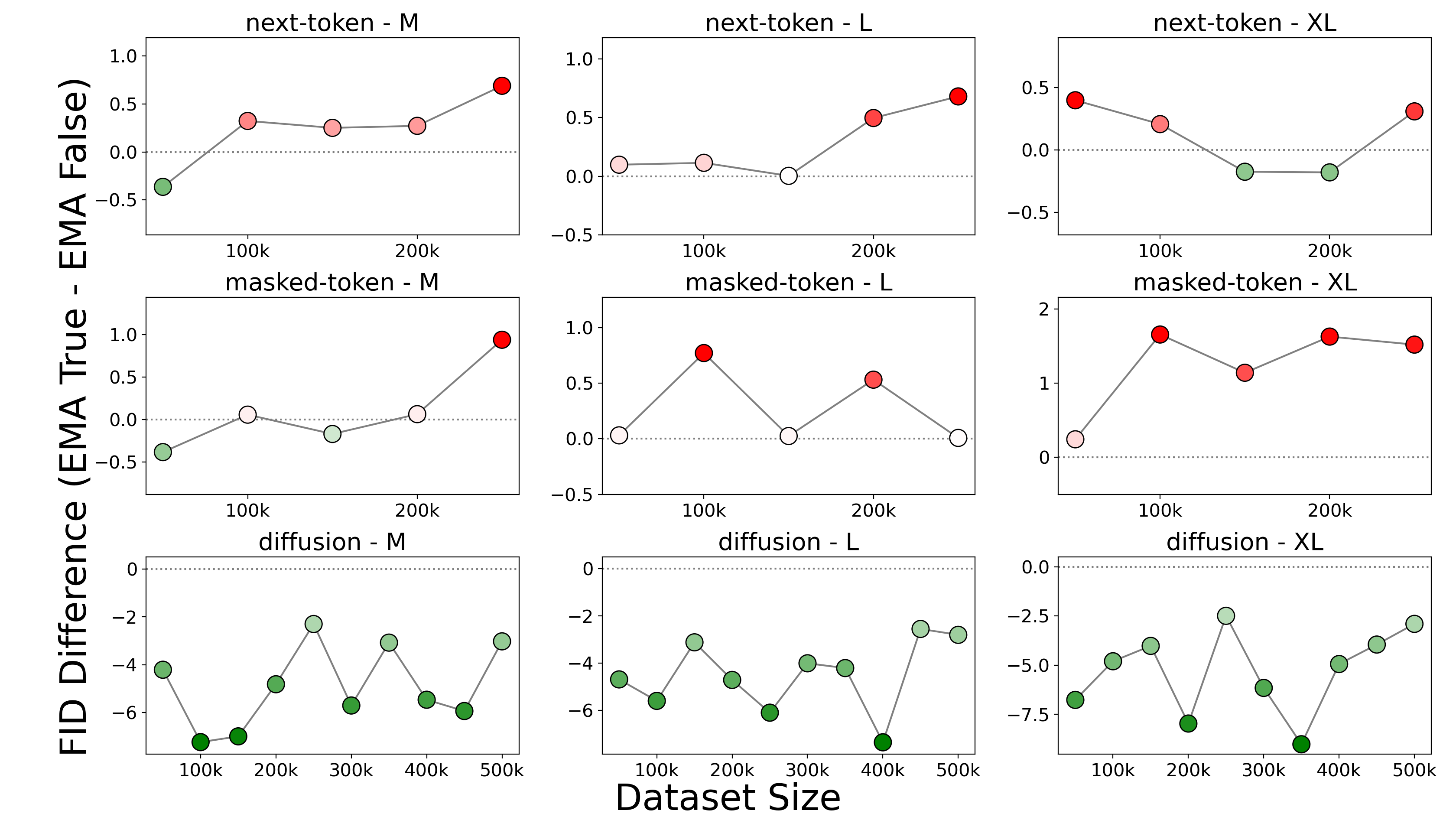}
    \label{fig:sd:fid}
\end{subfigure}
\caption{\textbf{Impact of EMA.} EMA significantly improves FID for diffusion models but hurts token based approaches. On CLIP score the effect on diffusion models stays consistent however for token based methods the influence is negligible. 
\vspace{-1em}}
\label{fig:emainfluence}
\end{figure*}
}

\newcommand{\infflops}{
\begin{figure*}[t]
\centering
\begin{subfigure}[t]{0.49\textwidth}
    \centering
    \includegraphics[width=0.9\linewidth]{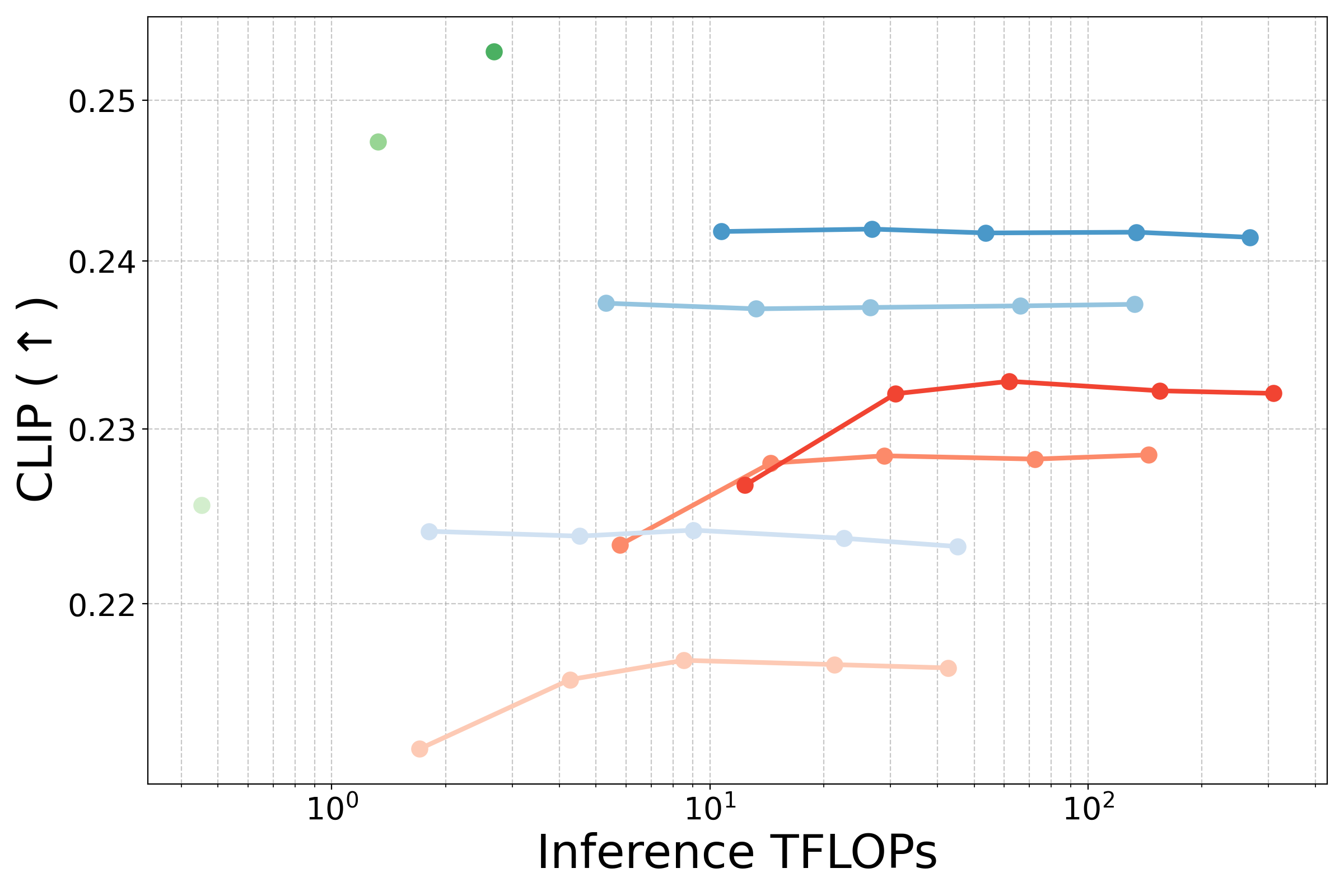}
    \label{fig:sdres:clip}
\end{subfigure}\hfill%
\begin{subfigure}[t]{0.49\textwidth}
    \centering
    \includegraphics[width=0.9\linewidth]{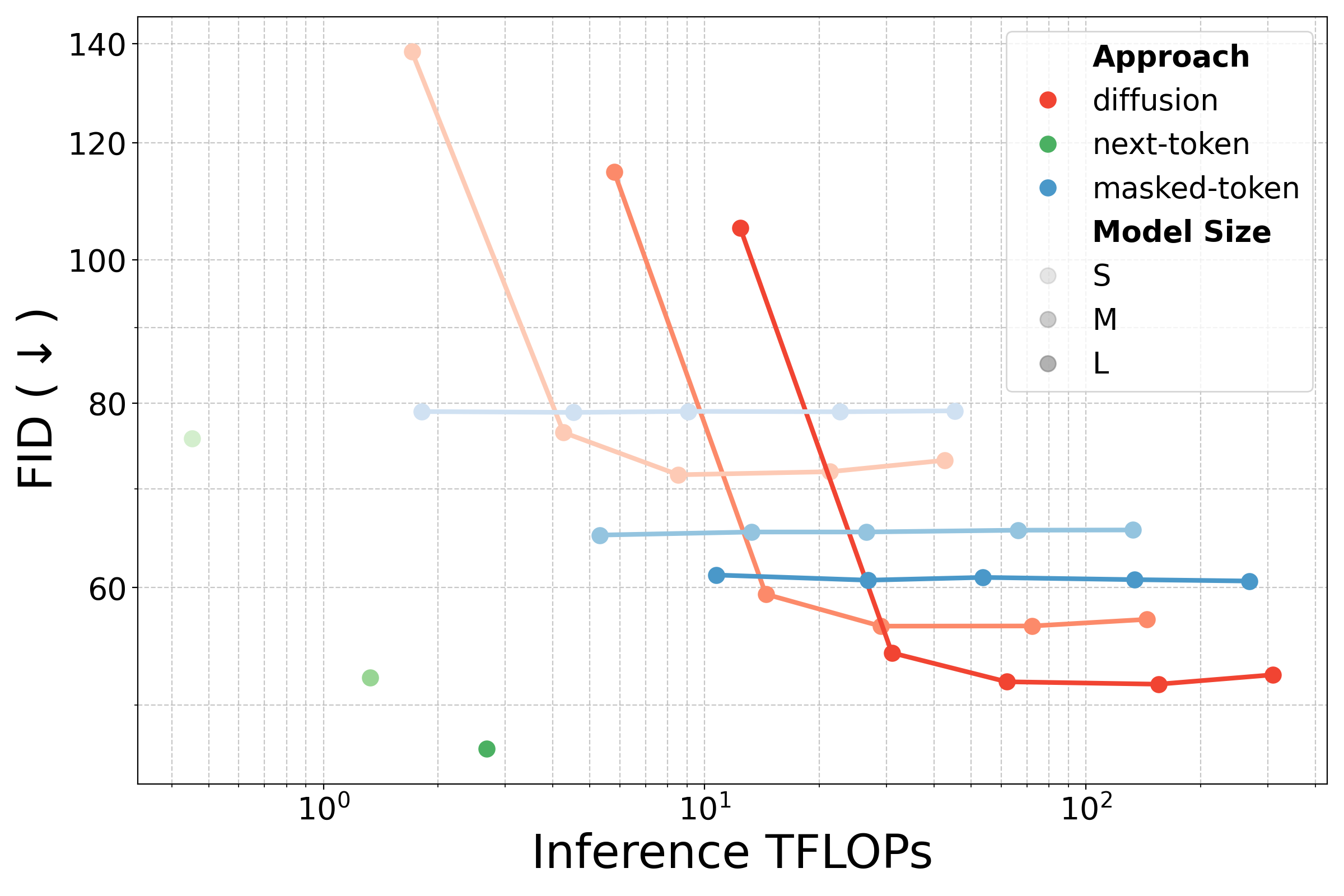}
    \label{fig:infflops:fid}
\end{subfigure}
\caption{\textbf{Inference compute efficiency on perceptual metrics.} Diffusion and masked token prediction evaluated at 4, 10, 20, 50, and 100 sampling steps. Next token prediction is 1 forward pass factorized over each token individually. Masked token prediction isn't influenced by the number of sampling steps very much. Next token prediction is the most compute efficient. Both axes are in log scale.
\vspace{-1em}}
\label{fig:infflops}
\end{figure*}
}

\newcommand{\scalingsamples}{
\begin{figure*}[t]
\centering
\includegraphics[width=0.75\linewidth]{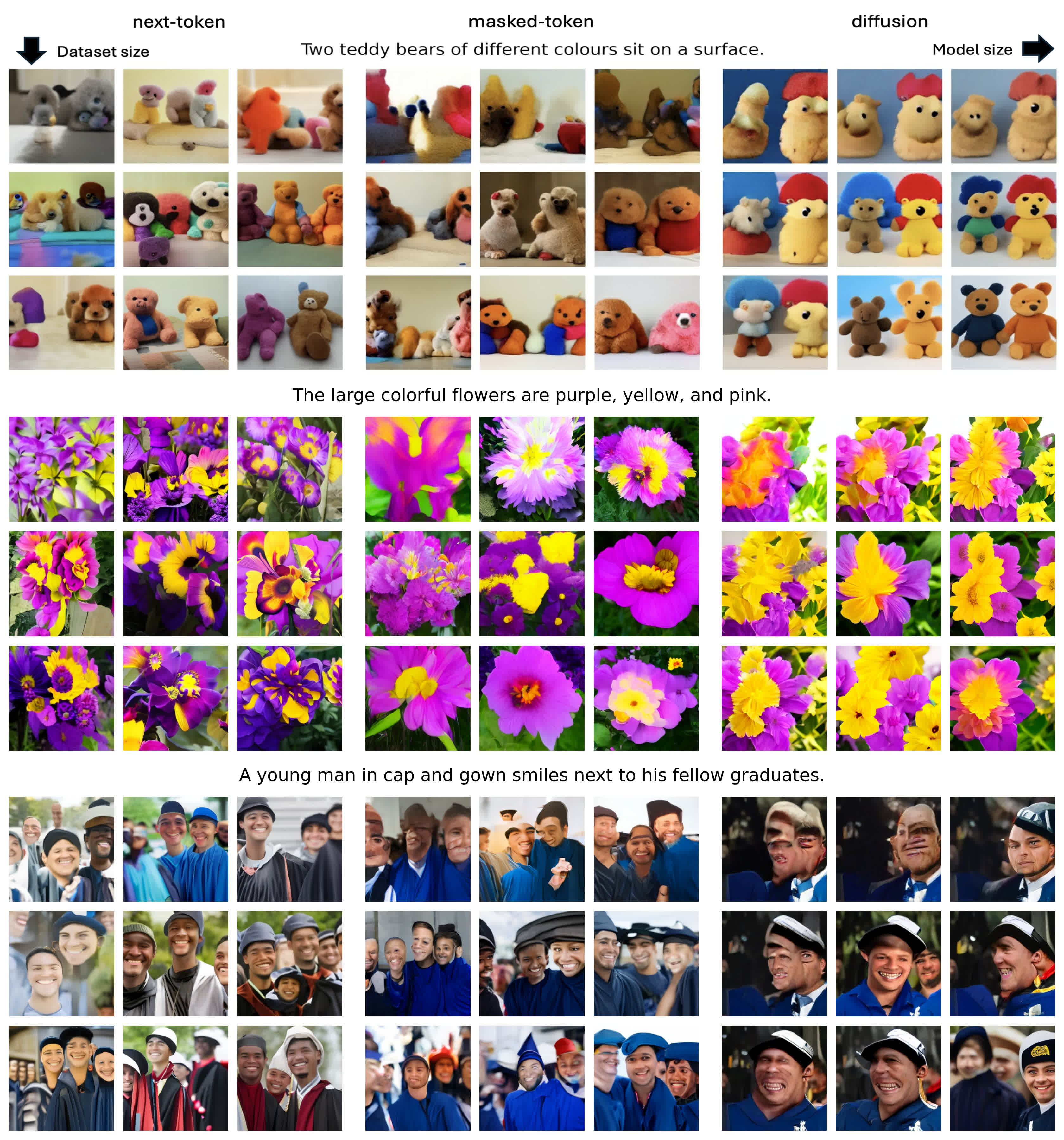}
\caption{\textbf{Increasing training compute improves sample quality for all approaches.} For each approach and prompt we sample an image with all combinations of S, M, L model sizes and 50k, 150k, 250k dataset sizes. Going down or right in the 3x3 increases dataset and model size respectively.
\vspace{-1em}}
\label{fig:scalingsamples}
\end{figure*}
}

\newcommand{\bestsamples}{
\begin{figure*}[t]
\centering
\includegraphics[width=1.0\linewidth]{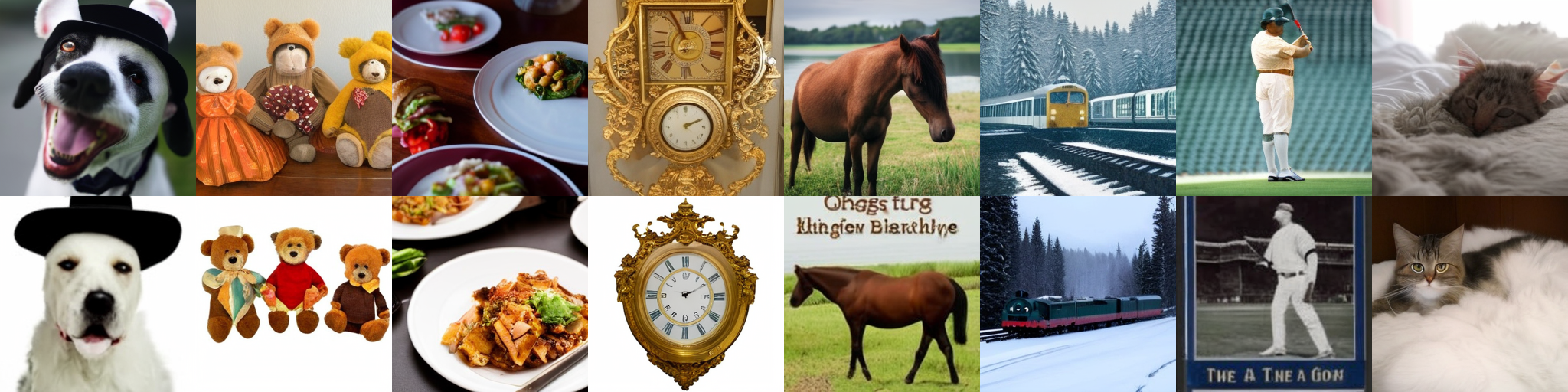}
\caption{\textbf{Images generated using our best models.} Top row is from a next-token prediction model, bottom row is from a diffusion model. Both models are XL size and trained for 500k steps.
\vspace{-1em}}
\label{fig:bestsamples}
\end{figure*}
}

\newcommand{\quantizers}{
\begin{figure*}[h!]
\centering

\begin{subfigure}{0.8\textwidth}
    \centering
    \includegraphics[width=\linewidth]{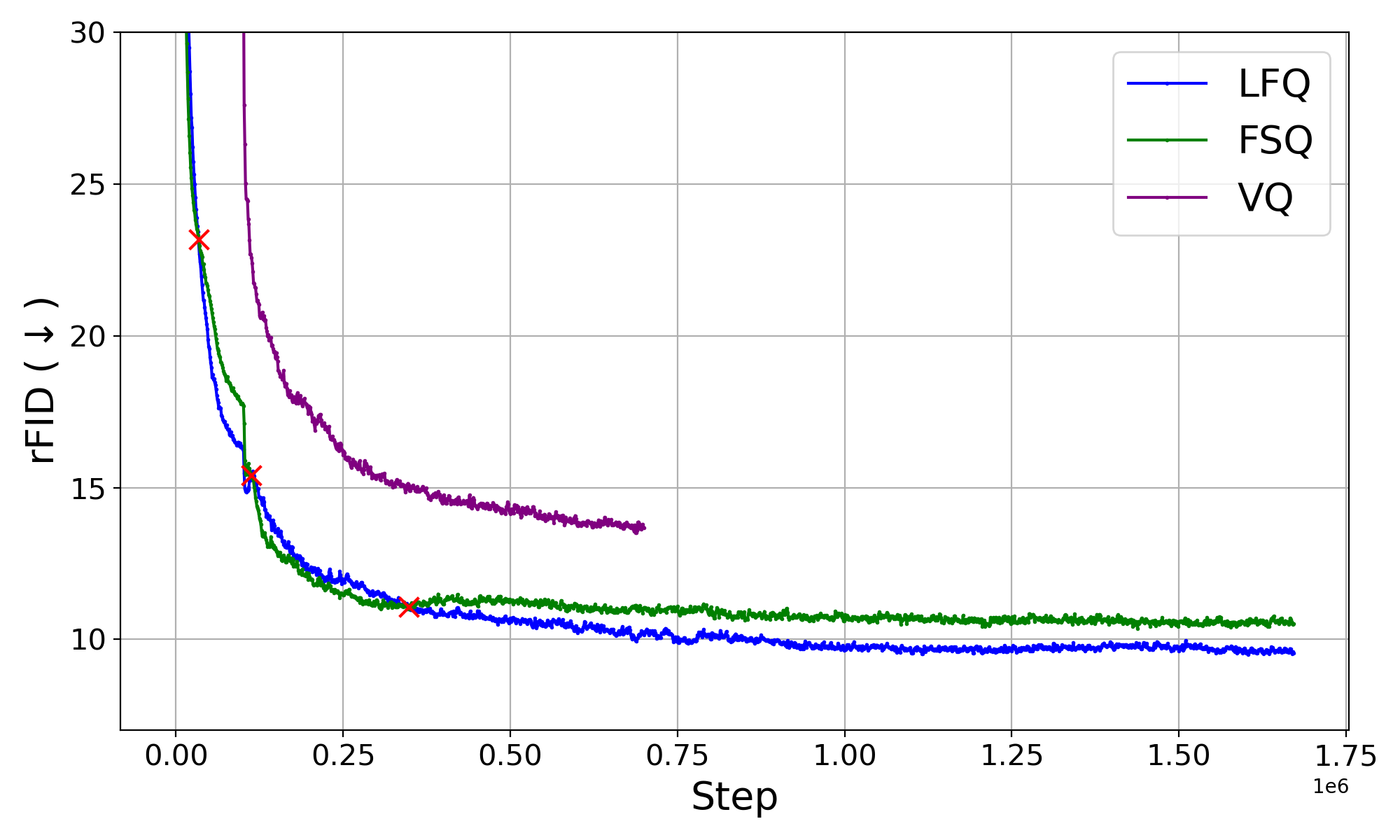}
\end{subfigure}\\[1ex] 

\begin{subfigure}{0.8\textwidth}
    \centering
    \includegraphics[width=\linewidth]{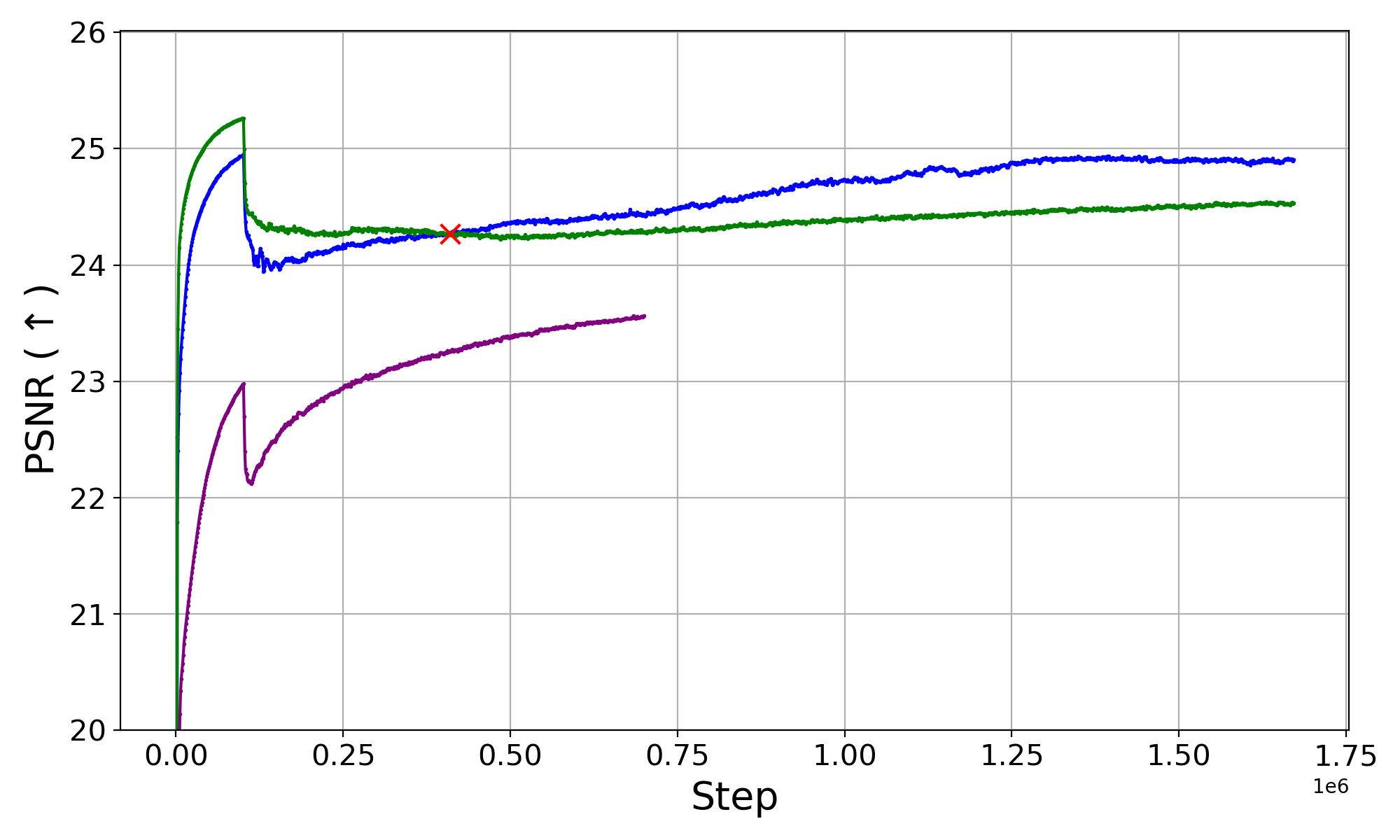}
\end{subfigure}\\[1ex] 

\begin{subfigure}{0.8\textwidth}
    \centering
    \includegraphics[width=\linewidth]{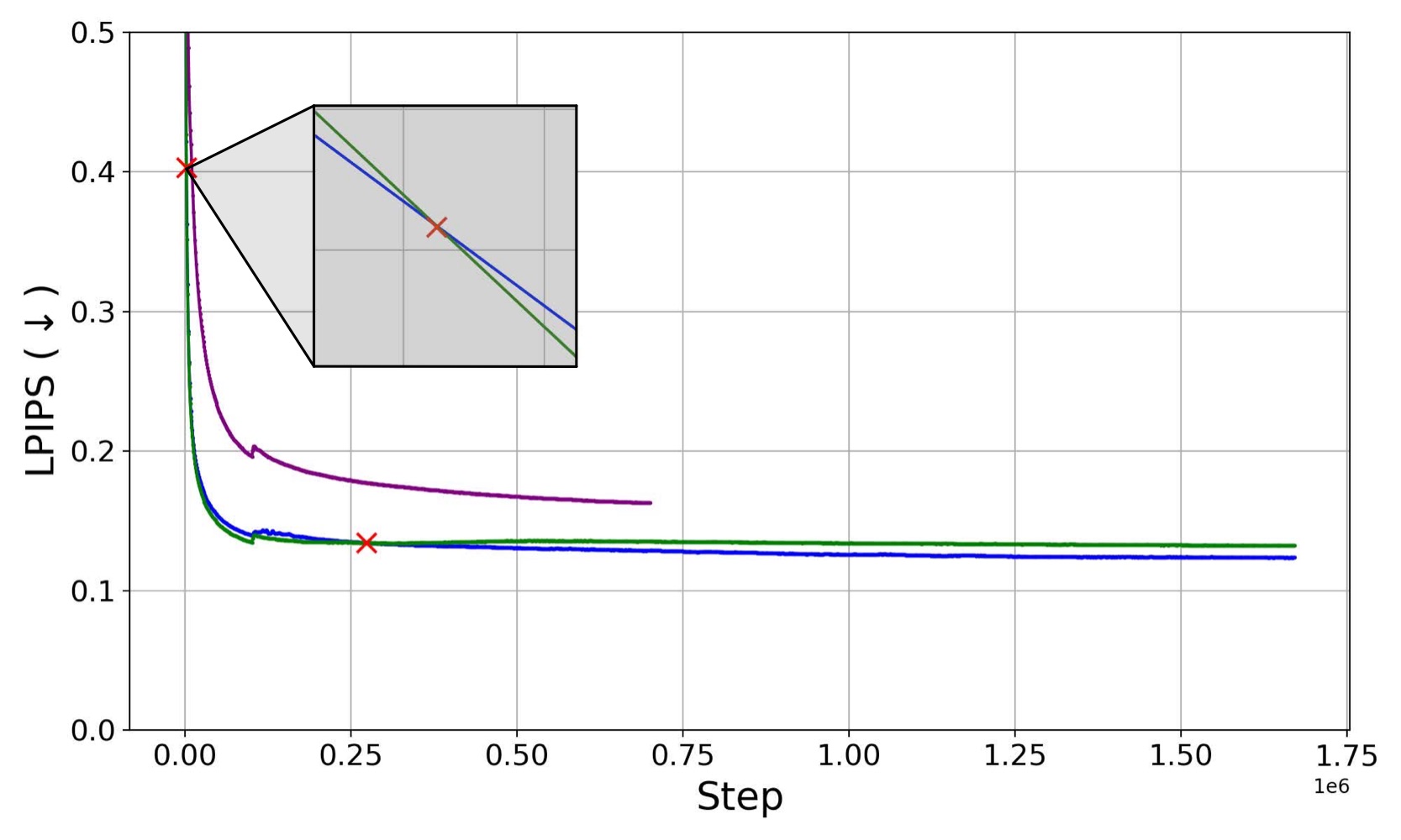}
\end{subfigure}

\caption{\textbf{Perceptual reconstruction metrics for various discrete regularization methods.} Classic vector quantization (VQ) struggles without tricks like codebook reinitialization. LFQ and FSQ have different training dynamics, often trading the lead in the beginning phases of training which is highlighted by the red X's.}
\vspace{-2ex} 
\label{fig:quantizers}
\end{figure*}
}
\begin{abstract}
\label{sec:abstract}
Nearly every recent image synthesis approach, including diffusion, masked-token prediction, and next-token prediction, uses a Transformer network architecture. Despite this common backbone, there has been no direct, compute controlled comparison of how these approaches affect performance and efficiency. We analyze the scalability of each approach through the lens of compute budget measured in FLOPs. We find that token prediction methods, led by next-token prediction, significantly outperform diffusion on prompt following. On image quality, while next-token prediction initially performs better, scaling trends suggest it is eventually matched by diffusion. We compare the inference compute efficiency of each approach and find that next token prediction is by far the most efficient. Based on our findings we recommend diffusion for applications targeting image quality and low latency; and next-token prediction when prompt following or throughput is more important.
\end{abstract}  
\section{Introduction}
\label{sec:introduction}

Following the work of \cite{peebles2023scalable}, deep image synthesis, including diffusion~\citep{sohl2015deep,song2019generative,song2020score,ho2020denoising,rombach2022highresolution,esser2024scaling}, masked-token prediction~\citep{chang2022maskgit, chang2023muse, villegas2022phenaki, yu2024language}, and next-token prediction~\citep{gafni2022makeascene, yu2022scaling, esser2021taming}, are all build on a common Transformer architecture \citep{vaswani2023attention}. 
Although these approaches are all known to scale well with compute and data, there has been relatively little controlled comparisons of their relative training and inference efficiency. Comparing these latent image synthesis approaches is challenging since the objectives they optimize often have different requirements which limit the set of applicable modules for each approach and influence their optimal configurations. For example, next-token prediction requires discrete input data which makes it unfit for continuous latent space regularization advances. In fact, latent image synthesis will be strongly influenced by the state of autoencoding research, often in an unbalanced way. Examples of this can be found in Section \ref{sec:relatedworks:latgenmod}.

In this paper, we measure the computational tradeoffs between popular transformer-based latent image synthesis approaches - diffusion, masked-token prediction, and next-token prediction. We investigate the impact of the autoencoder, which encodes the latent space, on generative results and train a large grid of models with the different approaches, model sizes, and dataset sizes.  Samples from some of our most capable models can be found in Figure \ref{fig:bestsamples}. Our findings indicate that (i) at smaller compute budgets, next-token prediction yields the best image quality but scaling trends suggest it is eventually matched by diffusion. (ii) Token-based approaches achieve superior controllability. (iii) The quality of the autoencoder impacts the FID more than the CLIP score of diffusion models trained on its latent space. (iv) We find preliminary evidence for improved diffusion training practices. Based on our findings, we recommend diffusion models for applications targeting low latency and high image quality; and next-token prediction for applications where prompt following and throughput are priorities.

\bestsamples
\section{Related Work}
\label{sec:relatedworks}

\textbf{Scaling transformer-based generative models}.
\label{sec:relatedworks:scaling}
Scaling compute budgets for transformer based generative models is a predictable method for improving performance. \cite{kaplan2020scaling, hoffmann2022training, clark2022unified} showed that for text, final training loss can be accurately predicted as a power law of training compute which depends on model size and dataset size. Following those practices many capable text generation models were trained \citep{touvron2023llama, brown2020language, rae2022scaling}. Similar results have been found for vision \citep{zhai2022scaling, alabdulmohsin2024getting, esser2024scaling, dehghani2023scaling} and even mixed modal data \citep{aghajanyan2023scaling}. We follow these intuitions and analyze image synthesis performance as a function of compute budget.

\textbf{Latent generative modeling}.
\label{sec:relatedworks:latgenmod}
Training latent generative vision models has emerged as an efficient alternative to the computationally intensive modeling of high-dimensional pixel space. Studies have demonstrated the advantages of imposing specific structural regularizations within the latent space for enhancing the performance of various generative models. For instance, \cite{rombach2022highresolution} observed that latent diffusion models operating in VAE-style \cite{kingma2022autoencoding} latent spaces, when regularized towards a standard Gaussian structure, outperform models trained with alternative regularization techniques. \cite{yu2024language, mentzer2023finite, yu2022vectorquantized} have shown that simplifying vector quantization methods can mitigate common issues such as poor codebook utilization and enhancing the transfer between autoencoder reconstruction quality and downstream generative model performance for token-based approaches. \cite{tian2024visual} demonstrated that employing hierarchical next-scale latents enables transformers using next token prediction to leverage their in-context learning capabilities more effectively, significantly improving performance. \cite{jin2024unified} use image latents dynamically sized based on their information content which allows generative models to allocate more computational resources to complex samples, as these will contain more tokens. We minimize potential bias coming from autoencoding asymmetries by studying the impact of the autoencoder on the generative model trained on top of it.

\section{Background}
\label{sec:methods}
\paragraph{Autoencoding}
To train latent generative models, we establish an encoder-decoder pair \((\mathcal{E}, \mathcal{D})\). For an image \(x \in \mathbb{R}^{H \times W \times 3}\), the encoder maps \(x\) to a latent representation \(z = \mathcal{E}(x)\), where \(z \in \mathbb{R}^{H/f \times W/f \times c}\) and \(f = 2^{m \in \mathbb{N}}\) represents the factor of dimensionality reduction. The decoder then reconstructs \(\hat{x} = \mathcal{D}(z)\), aiming for high perceptual similarity to \(x\), effectively making \(z\) a perceptually compressed representation of the input. To avoid high-variance latent spaces and ensure structured representations, we employ regularization methods classified into two main types: discrete and continuous. The regularization function \(\textbf{q}\) for a continuous regularizer maps \(\mathbb{R}^d \to \mathbb{R}^d\), while a discrete regularizer maps \(\textbf{q}: \mathbb{R}^d \to \{0, 1, 2, \ldots, N\}\), making the latent space finite. In the following subsections we use $z \in \mathbb{R}^{s \times d}$ to denote the flattened representation output by the encoder $\mathcal{E}$. $p(z)$ is the latent data distribution we are interested in estimating using our generative models. We recover images by inputting sampled latents into the corresponding decoder $\mathcal{D}$.

\paragraph{Next token prediction}
In the context of sequences of discrete tokens represented as \( z \in \{0, 1, 2, \ldots, N\}^{s} \), we employ the chain rule of conditional probability to decompose the target distribution into a product of conditional distributions which are tractable since the range of $z_i$ is finite. To model this distribution, we use a neural network \( f \), parameterized by weights \( \theta \). The parameters are optimized by minimizing the negative log-likelihood \( \mathcal{L}_{NT} \).
\vspace{-2pt}
\begin{align}
p(z) &= \prod_{i=1}^{n} p(z_i | z_{i-1}, \ldots, z_1) & \mathcal{L}_{NT} &= \mathbb{E}_i [ -\log{p(z_i | z_{<i}; \theta}) ]
\end{align}
\vspace{-2pt}
Sampling from our learned distribution begins with an empty sequence (in practice, a "start of text" token is sampled with 1.0 probability). We then sample the first token unconditionally and append it to our sequence. The process continues by iteratively evaluating the conditionals and sampling from them, with each step increasing the sequence length by one.

\paragraph{Masked token prediction}
Masked token prediction is a form of iterative denoising and can be viewed as a discrete diffusion process. In this process, tokens progressively transition to an absorbing [MASK] state according to a probability defined by a noise schedule \(\gamma(t) \in (0, 1]\) where $t \sim \mathcal{U}(0, 1)$. This transition can also be mathematically expressed as a product of conditionals, except in a perturbed order $\sigma$, and implemented as a neural network. Here, \(\sigma(i)\) is a surjective function mapping \([0, N] \mapsto [0, N]\). We follow \cite{chang2022maskgit, chang2023muse} where $\sigma(i) = \sigma(i, t)$ such that \(p(\sigma(i, t) < j) = \gamma(t)\) meaning the likelihood a token can be attended to is independent of position. In this formulation, we utilize a truncated \( \arccos \) distribution for our noise schedule: $\gamma(t) = \frac{2}{\pi} (1 - t^2)^{-\frac{1}{2}}$. To apply this method, we generate a mask tensor \(M \in \{0, 1\}^s\) by sampling \(t \sim \mathcal{U}(0, 1)\) and \(m_i \sim \text{Bernoulli}(\gamma(t))\). The tensor \(M\) is applied elementwise to the latents, replacing \(z_i\) with the [MASK] token if \(m_i = 1\); otherwise, \(z_i\) remains unchanged. Denote the resultant noised sequence as \(z_M\). The network is then trained to minimize the masked token loss \(\mathcal{L}_{MT}\).
\vspace{-2pt}
\begin{align}
p(z) &= \prod_{i, \sigma(i) = j}^{n} p(z_j | z_{\sigma(i) < j}) & \quad \mathcal{L}_{MT} &= \mathbb{E}_{i, m_i=1} [ -\log{p(z_i | z_{\overline{M}}; \theta}) ]
\end{align}
\vspace{-2pt}
Sampling from the distribution starts with a fully masked sequence and iterates through a discretized noise schedule \(t_i = i/N\) over \(N\) desired steps. At each step, the model estimates \(p(z | z_{\overline{M}})\) for sampling, followed by re-noising using \(\gamma(t_{i+1})\). This iterative re-noising and sampling process is repeated \(N\) times to yield the final sample.

\paragraph{Diffusion}
We adopt the flow matching framework outlined by \cite{lipman2023flow}, focusing on models that map samples from a noise distribution \( p_1 \) to a data distribution \( p_0 \) using continuous trajectories governed by an ordinary differential equation (ODE). Furthermore, we enforce straight paths between the terminal distributions (by setting \( \alpha_t = 1 - t \) and \( \beta_t = t \))  since this has been shown to perform well at scale \citep{esser2024scaling}.
\vspace{-2pt}
\begin{align}
\label{eq:ode}
d\phi_t(x) &= v_t(\phi_t(x)) \, dt & \phi_0(x) &= x & z_t &= \alpha_t x_0 + \beta_t \epsilon, \quad \epsilon \sim \mathcal{N}(0, 1)
\end{align}
\vspace{-2pt}
Here, \( v_t : [0, 1] \times \mathbb{R}^d \mapsto \mathbb{R}^d \) represents a time-dependent vector field, which we aim to parameterize using a neural network \( \theta \) and $\phi_t : [0, 1] \times \mathbb{R}^d \mapsto \mathbb{R}^d$ is the flow. To optimize our weights, we regress the vector field \( u_t \), which generates our paths \( z_t \) by employing conditional flow matching which we reformulate as a noise-prediction objective $\mathcal{L}_{CFM}$. Sampling is performed by using an ODE solver to solve Equation \ref{eq:ode} in reverse time, utilizing our trained neural network \( v_{\Theta}(z, t) \).

\begin{equation*}
  \mathcal{L}_{CFM}(x_0) = \mathbb{E}_{t \sim \mathcal{U}(t), \epsilon \sim \mathcal{N}(0, I)}
  \left[ \frac{1}{(1-t)^2} \Vert \epsilon_\theta(z_t, t) - \epsilon \Vert^2 \right]\;
\end{equation*}
\section{Experimental Setup}
\label{sec:experimentalsetup}

\paragraph{Data}
We train both the autoencoders and the generative models on a large web dataset of image and text pairs at 256x256 resolution. For the conditioning we use the pooled text embedding of the OpenCLIP bigG/14 model from \cite{cherti2022reproducible}. Once the autoencoders are trained we pre-encode the entire dataset with them for improved training speed.

\paragraph{Evaluation metrics}
\label{sec:experimentalsetup:eval}
Since we are in the infinite data regime, we look at the final train loss and do not compare across objectives since the losses represent different quantities. We also look at CLIP score \citep{radford2021learning, hessel2022clipscore} and FID computed on CLIP features \citep{sauer2021projected} based on the decoded samples $\hat{x} = \mathcal{D}(z)$.

\paragraph{Autoencoding}
\label{experimentalsetup:autoencoding}
We study well-established autoencoder configurations that have proven effective without special handling for each data type. We adhere to the training and architectural guidelines provided by \cite{rombach2022highresolution}. Each autoencoder is trained with a downsampling factor \(f=8\), reducing \(256 \times 256\) images to a \(32 \times 32\) grid of latents. For continuous variants, \(\textbf{q}(z)\) implements a KL penalty aiming towards the standard normal distribution \citep{kingma2022autoencoding, rombach2022highresolution}, while for discrete variants, we utilize lookup-free quantization (LFQ) \citep{yu2024language}. Further details on the selection of discrete regularizers are available in Appendix \ref{appendix:discrete_regularizers}. To circumvent potential challenges associated with large vocabulary sizes, as highlighted by \cite{yu2024language}, our LFQ-regularized autoencoder is trained with a vocabulary size of 16384~\citep{esser2021taming}. Assessing the comparability of autoencoders is difficult since there are many variables of interest such as the (1) information capacity of the latent space; (2) compute used to train the autoencoder; (3) reconstruction quality achieved by the autoencoder. To explore the influence of these factors on the performance of generative models, we train a set of autoencoders similar to those in \cite{esser2024scaling}, which exhibit a range of information capacities and reconstruction qualities. Additionally, we experiment with targeting specific reconstruction qualities, irrespective of other factors, by training a KL-regularized autoencoder with early stopping to match the reconstruction quality of our discrete autoencoder within a certain threshold \(\epsilon\) \footnote{Discrete autoencoders typically have worse reconstruction qualities since the information bottleneck is tighter. This can be shown by comparing $\log(\text{codebook size})$ to $\text{num\_channels} * \text{sizeof(dtype)}$ for common values of these quantities. In our case we needed to stop at 75k steps vs. 1M for the discrete autoencoder.}. Table \ref{tab:aereconstructiontable} provides detailed information about the autoencoders.

\aereconstructiontable

\textbf{Autoencoder ablation.} We train an L-size diffusion model on top of the latent space of each continuous autoencoder. We then evaluate the models using the metrics described in Section \ref{sec:experimentalsetup:eval} and plot them against the number of training steps. Results are shown in Figure \ref{fig:diffae}. We find that the autoencoder's reconstruction quality has a consistently significant impact on the FID score, while its effect on the CLIP score diminishes with larger dataset sizes, where the models tend to yield similar results. This trend likely emerges because improvements in autoencoder quality enhance perceptual reconstruction metrics similar to FID, rather than affecting language or semantic capabilities. Upon examining the number of channels in the autoencoders, our findings concur with those reported by \cite{esser2024scaling}, indicating that leveraging larger and better latent spaces requires more compute and model capacity. Additionally, the model trained on our early-stopped autoencoder's latent space performed significantly worse than the 4-channel autoencoder, which achieves similar reconstruction quality. This confirms the importance of latent space structure for overall performance.
Building on these insights, we have chosen to use the 4-channel autoencoder for our main diffusion experiments. This model most closely matches the latent space capacity and reconstruction quality of our discrete autoencoder, while also ensuring that the latent structure is adequately developed to support the diffusion model trained on it. Although more advanced autoencoders have been developed—such as those featuring increased channel counts or expanded codebook sizes—our primary focus in this study is to maintain comparability across objectives.

\diffae

\subsection{Network Architecture}
\label{experimentalsetup:networkarch}
\textbf{Backbone.} We opt for the transformer architecture as our primary network backbone, recognizing its capability to scale effectively with computational resources and its status as the state-of-the-art (SOTA) across all evaluated approaches. Configuring a transformer involves many decisions, such as choosing normalization methods, feed-forward layer configurations, positional embedding schemes, conditioning methods, and initialization strategies. Given the prohibitive cost of exploring all possible hyperparameters, we adhere to established practices in recent studies.

\textbf{Design differences.} For approaches utilizing discrete representations, we primarily follow the configurations used in the LLaMa model \citep{touvron2023llama}, incorporating SwiGLU feed-forward layers with an expansion ratio of \( \frac{2}{3}4 \) and rotary positional embeddings \citep{su2023roformer}. An exception is made for masked token prediction, where learned positional embeddings are preferred to address positional ambiguities that degrade performance near the center of the image. For continuous representation approaches, we align with diffusion transformers \citep{peebles2023scalable}, employing GELU feed-forward layers with an expansion ratio of 4 and learned positional embeddings. All models use QK-normalization \citep{dehghani2023scaling} for better training stability.

\textbf{Conditioning ablation.} We choose to ablate the conditioning method, as it significantly impacts the computational cost of model operations. Adaptive layer normalization (AdaLN) \citep{perez2017film} has shown promise in latent image synthesis for both continuous \citep{peebles2023scalable} and discrete \citep{tian2024visual} settings. To validate this choice in the discrete context, we conduct small-scale ablations on S-size models, comparing AdaLNzero \citep{peebles2023scalable}, with two other common conditioning methods: prepending a projected embedding in the context of the transformer and cross-attention. The outcomes of these ablations are presented in Table \ref{tab:condablation}, informing our choice of conditioning method for subsequent experiments.


\textbf{Compute cost.} To assess the computational cost of each model, we first standardize a set of hyperparameters across all transformers, detailed in Table \ref{tab:tfsizetable}. We then calculate the forward pass FLOPs for a single sample (a sequence of 1024 embeddings) for each approach and model size, and present them in Table \ref{tab:modelflopstable}. Assuming the backward pass is twice the cost of the forward pass, we compute the training FLOPs for each model as \( (1 + 2) \times (\text{forward FLOPs}) \times D \), where \( D \) represents the total number of training samples.

\modelflopstable

\subsection{Training}
Each approach also has associated training hyperparameters which past work has found to work well and for the same reasons as stated in \ref{experimentalsetup:networkarch} we follow them. 

\textbf{Optimization and conditioning.} For diffusion experiments we follow \cite{esser2024scaling} and use a constant learning rate schedule with a maximum value of $1^{-4}$. For next and masked token prediction we use a cosine decay learning rate with a maximum value of $3^{-3}$ which decays down to $3^{-5}$. All models have a linear learning rate warmup lasting 1000 steps up to the maximum value. We use the AdamW \citep{loshchilov2019decoupled} optimizer with $\beta_1 =0.9$, $\beta_2 = 0.95$, decay=0.01, and epsilon=1e-15 for improved transformer training stability \citep{wortsman2023smallscale}. All models are trained at bf16-mixed precision \citep{bfloat16}. We intend to use classifier free guidance (CFG) \citep{ho2022classifierfree} during sampling so we randomly drop conditioning 10\% of the time during training. Since its inexpensive and does not influence training, for all models, we store a copy of the model weights which gets updated every 100 training batches with an exponential moving average (EMA) using a decay factor of 0.99 and during evaluation we evaluate both sets of weights. 

\textbf{Training steps.} For each objective and model size we scale to at least 250k training steps with a batch size of 512. For diffusion we decide to go up to 500k steps since constant learning rate schedule allows more flexibility with dataset size\footnote{With a decaying learning rate, each dataset size we want to study requires a separate run from scratch whereas for constant learning rate schedules you can simply continue from a past checkpoint}. Occasionally we train models for longer to attempt to illustrate convergence or crossing points.

\subsection{Sampling}
\textbf{Classifier free guidance.} \cite{ho2022classifierfree} introduced it in diffusion models as an elegant way of trading off diversity for fidelity and has been demonstrated to improve results for all approaches we consider in this study \citep{chang2023muse, gafni2022makeascene, ho2022classifierfree}. We use it here in the form 
\begin{equation}
x_g = (1 + w)x_c - wx_u
\end{equation}
where $w$ is the guidance scale. For diffusion $x$ will be the position in the denoising trajectory and for token based methods $x$ is the logit distribution at a given timestep.

\textbf{Hyperparameters.} For our diffusion models we follow \cite{esser2024scaling} and use 50 sampling steps with a CFG scale of 5. Since the conditioning and input data is slightly different we also perform a small sweep around those parameters to confirm they are still optimal. For the token based models we could not find good resources on reasonable sampling hyperparameters so we perform small sweeps for S-size models to find the best configurations and verify the robustness of those values for larger models. Common between them, we use nucleus sampling \citep{holtzman2020curious} with a top-p value of $0.9$ and a temperature of $1.0$. For next token prediction and masked token prediction we use CFG scales 8 and 5 respectively. For masked token prediction we perform 10 sampling steps.
\section{Results}
\label{sec:results}
\subsection{Training tradeoffs}

For all models, we measure our evaluation metrics every 50k steps of training and plot them in log scale against the log of training compute. Figure \ref{fig:main} presents this for FID and CLIP score. There we can see that for FID, next token prediction starts out more compute efficient but scaling trends suggest that its eventually matched by diffusion. When looking at CLIP score we see that token prediction is significantly better than diffusion, implying the models generate images that follow the input prompt better. This could be a feature of using more compressed latent spaces which is supported by Figure \ref{fig:diffae} where the 4 channel continuous autoencoder outperforms both the 8 and 16 channel autoencoder on CLIP score near the end of training.  This is also supported in Figure \ref{fig:scalingsamples} with interpretable features like human faces emerging sooner in the token based methods. Extending a finding from \cite{mei2024bigger}, we observe that, for all approaches studied, smaller models trained for longer often surpass larger models. In Figure \ref{fig:fintl} we show the final training loss of each model against training compute to show that it follow similar scaling trends to what has been shown in past work on scaling deep neural networks, briefly described in Section \ref{sec:relatedworks:scaling}. Samples from the most capable XL sized next-token prediction and diffusion models can be found in Figure \ref{fig:bestsamples}.

\mainres
\fintl

\subsection{Inference tradeoffs}
\label{sec:results:inference}
\textbf{Inference cost.} We evaluated all models trained for 250k steps to understand the impact of inference FLOPs on perceptual metrics. To adjust the number of inference FLOPs for a single model, we varied the number of sampling steps, applicable only to iterative denoising methods like masked token prediction and diffusion. As shown in Figure \ref{fig:infflops}, next-token prediction demonstrates far greater inference compute efficiency compared to other objectives. This efficiency arises because when using key-value caching, sampling N tokens autoregressively uses the same amount of FLOPs as forwarding those N tokens in parallel once. However, for iterative denoising methods, this value is multiplied by the number of sampling steps. Interestingly, despite being trained for iterative denoising, the number of steps in masked token prediction appears to have minimal impact on sample quality.

\infflops

\textbf{Sampling latency and throughput.} While next-token prediction requires much less compute per sample, the autoregressive dependency of each token causes it to be data bound when few queries are being processed in parallel which results in high latency. Conversely, bidirectional denoising approaches utilize a more parallel sampling process which, despite its high cost, facilitates low latency especially in low-volume settings with models that fit on local devices \citep{chang2022maskgit}. For high-volume sampling, where throughput becomes more important, such as serving many users via an API, next token prediction could use a batching algorithm to maximize GPU utilization by choosing batch sizes inversely proportional to sequence lengths. The effectiveness of this method is ensured by the fact that, for next-token prediction image synthesis, all responses are the same length so you can easily plan your batches ahead. This way, for high-volume sampling, next-token prediction would enjoy the same benefits over the other approaches as presented in the cost section above but for sample throughput.

\subsection{EMA ablations}
Among the various training practices distinguishing these methods, the use of an exponential moving average (EMA) on the model weights stands out. In the diffusion literature \citep{karras2024analyzing, karras2022elucidating, peebles2023scalable, esser2024scaling} EMA is an essential component of the training pipeline. In contrast, this practice has not received equivalent attention in other approaches. The differential impact of EMA is evident in Figure \ref{fig:emainfluence}. For token-based approaches, the influence of EMA is either negligible or, in some cases, harmful, whereas for diffusion models, it is beneficial almost universally. We hypothesize that the impact of EMA may be linked to the learning rate schedule, where decaying schedules similarly minimize weight variation towards the end of training. To test this hypothesis, we conducted an ablation study on an M-sized next-token prediction and diffusion model trained over 250k steps. Our findings verify our hypothesis that EMA enhances performance under a constant learning rate schedule; however, it does not exceed the improvements seen with a cosine decay learning rate schedule. This implies that future diffusion models should consider substituting the EMA for a cosine decay learning rate schedule if they are willing to pay the cost of decreased training length flexibility. Results from this ablation study are presented in Table \ref{tab:emaablation}. 

\emainfluence
\emaablation

\subsection{Limitations}
\label{sec:limitations}

Our analysis has several limitations which result from resource limitations and project scope. We only investigate pretraining whereas most production systems utilize a progression of pretraining, finetuning, and distillation stages. We do not investigate high resolution images. We only measure loss and perceptual metrics and leave out an analysis of utility for potential downstream tasks. There are many others approaches that we leave out such as other discrete diffusion approaches \citep{austin2023structured, pernias2023wuerstchen}, causally masked token prediction \citep{aghajanyan2022cm3}, and many more. We choose most hyperparameters by following past work instead of exhaustively sweeping to find the best configurations. And finally, we do not compare approaches using the best possible autoencoders.

\scalingsamples
\section{Conclusion}
\label{sec:conclusion}

We conduct a compute-controlled analysis comparing transformer-based diffusion, next-token prediction, and masked-token prediction latent image synthesis models. Our findings indicate that token based methods, led by next-token prediction, achieve superior CLIP scores, indicating greater controllability. In terms of FID, and therefore image quality, while next-token prediction is much better at low training compute scales, scaling trends suggest it is eventually matched by diffusion. We find that next token prediction has, by far, the best inference compute efficiency but this comes at the cost of high latency in low data intensity settings. Based on our findings recommend diffusion models when image quality and low latency is important; and next-token prediction for better prompt following and throughput.

\bibliographystyle{plainnat}
\bibliography{references}

\begin{thebibliography}{49}
\providecommand{\natexlab}[1]{#1}
\providecommand{\url}[1]{\texttt{#1}}
\expandafter\ifx\csname urlstyle\endcsname\relax
  \providecommand{\doi}[1]{doi: #1}\else
  \providecommand{\doi}{doi: \begingroup \urlstyle{rm}\Url}\fi

\bibitem[Aghajanyan et~al.(2022)Aghajanyan, Huang, Ross, Karpukhin, Xu, Goyal, Okhonko, Joshi, Ghosh, Lewis, and Zettlemoyer]{aghajanyan2022cm3}
Armen Aghajanyan, Bernie Huang, Candace Ross, Vladimir Karpukhin, Hu~Xu, Naman Goyal, Dmytro Okhonko, Mandar Joshi, Gargi Ghosh, Mike Lewis, and Luke Zettlemoyer.
\newblock Cm3: A causal masked multimodal model of the internet, 2022.

\bibitem[Aghajanyan et~al.(2023)Aghajanyan, Yu, Conneau, Hsu, Hambardzumyan, Zhang, Roller, Goyal, Levy, and Zettlemoyer]{aghajanyan2023scaling}
Armen Aghajanyan, Lili Yu, Alexis Conneau, Wei-Ning Hsu, Karen Hambardzumyan, Susan Zhang, Stephen Roller, Naman Goyal, Omer Levy, and Luke Zettlemoyer.
\newblock Scaling laws for generative mixed-modal language models, 2023.

\bibitem[Alabdulmohsin et~al.(2024)Alabdulmohsin, Zhai, Kolesnikov, and Beyer]{alabdulmohsin2024getting}
Ibrahim Alabdulmohsin, Xiaohua Zhai, Alexander Kolesnikov, and Lucas Beyer.
\newblock Getting vit in shape: Scaling laws for compute-optimal model design, 2024.

\bibitem[Austin et~al.(2023)Austin, Johnson, Ho, Tarlow, and van~den Berg]{austin2023structured}
Jacob Austin, Daniel~D. Johnson, Jonathan Ho, Daniel Tarlow, and Rianne van~den Berg.
\newblock Structured denoising diffusion models in discrete state-spaces, 2023.

\bibitem[Brown et~al.(2020)Brown, Mann, Ryder, Subbiah, Kaplan, Dhariwal, Neelakantan, Shyam, Sastry, Askell, Agarwal, Herbert-Voss, Krueger, Henighan, Child, Ramesh, Ziegler, Wu, Winter, Hesse, Chen, Sigler, Litwin, Gray, Chess, Clark, Berner, McCandlish, Radford, Sutskever, and Amodei]{brown2020language}
Tom~B. Brown, Benjamin Mann, Nick Ryder, Melanie Subbiah, Jared Kaplan, Prafulla Dhariwal, Arvind Neelakantan, Pranav Shyam, Girish Sastry, Amanda Askell, Sandhini Agarwal, Ariel Herbert-Voss, Gretchen Krueger, Tom Henighan, Rewon Child, Aditya Ramesh, Daniel~M. Ziegler, Jeffrey Wu, Clemens Winter, Christopher Hesse, Mark Chen, Eric Sigler, Mateusz Litwin, Scott Gray, Benjamin Chess, Jack Clark, Christopher Berner, Sam McCandlish, Alec Radford, Ilya Sutskever, and Dario Amodei.
\newblock Language models are few-shot learners, 2020.

\bibitem[Chang et~al.(2022)Chang, Zhang, Jiang, Liu, and Freeman]{chang2022maskgit}
Huiwen Chang, Han Zhang, Lu~Jiang, Ce~Liu, and William~T. Freeman.
\newblock Maskgit: Masked generative image transformer, 2022.

\bibitem[Chang et~al.(2023)Chang, Zhang, Barber, Maschinot, Lezama, Jiang, Yang, Murphy, Freeman, Rubinstein, Li, and Krishnan]{chang2023muse}
Huiwen Chang, Han Zhang, Jarred Barber, AJ~Maschinot, Jose Lezama, Lu~Jiang, Ming-Hsuan Yang, Kevin Murphy, William~T. Freeman, Michael Rubinstein, Yuanzhen Li, and Dilip Krishnan.
\newblock Muse: Text-to-image generation via masked generative transformers, 2023.

\bibitem[Chen et~al.(2019)Chen, Chou, Xu, and Hseu]{bfloat16}
Dehao Chen, Chiachen Chou, Yuanzhong Xu, and Jonathan Hseu.
\newblock Bfloat16: The secret to high performance on cloud tpus, 2019.
\newblock URL \url{https://cloud.google.com/blog/products/ai-machine-learning/bfloat16-the-secret-to-high-performance-on-cloud-tpus?hl=en}.

\bibitem[Cherti et~al.(2022)Cherti, Beaumont, Wightman, Wortsman, Ilharco, Gordon, Schuhmann, Schmidt, and Jitsev]{cherti2022reproducible}
Mehdi Cherti, Romain Beaumont, Ross Wightman, Mitchell Wortsman, Gabriel Ilharco, Cade Gordon, Christoph Schuhmann, Ludwig Schmidt, and Jenia Jitsev.
\newblock Reproducible scaling laws for contrastive language-image learning, 2022.

\bibitem[Clark et~al.(2022)Clark, de~las Casas, Guy, Mensch, Paganini, Hoffmann, Damoc, Hechtman, Cai, Borgeaud, van~den Driessche, Rutherford, Hennigan, Johnson, Millican, Cassirer, Jones, Buchatskaya, Budden, Sifre, Osindero, Vinyals, Rae, Elsen, Kavukcuoglu, and Simonyan]{clark2022unified}
Aidan Clark, Diego de~las Casas, Aurelia Guy, Arthur Mensch, Michela Paganini, Jordan Hoffmann, Bogdan Damoc, Blake Hechtman, Trevor Cai, Sebastian Borgeaud, George van~den Driessche, Eliza Rutherford, Tom Hennigan, Matthew Johnson, Katie Millican, Albin Cassirer, Chris Jones, Elena Buchatskaya, David Budden, Laurent Sifre, Simon Osindero, Oriol Vinyals, Jack Rae, Erich Elsen, Koray Kavukcuoglu, and Karen Simonyan.
\newblock Unified scaling laws for routed language models, 2022.

\bibitem[Dehghani et~al.(2023)Dehghani, Djolonga, Mustafa, Padlewski, Heek, Gilmer, Steiner, Caron, Geirhos, Alabdulmohsin, Jenatton, Beyer, Tschannen, Arnab, Wang, Riquelme, Minderer, Puigcerver, Evci, Kumar, van Steenkiste, Elsayed, Mahendran, Yu, Oliver, Huot, Bastings, Collier, Gritsenko, Birodkar, Vasconcelos, Tay, Mensink, Kolesnikov, Pavetić, Tran, Kipf, Lučić, Zhai, Keysers, Harmsen, and Houlsby]{dehghani2023scaling}
Mostafa Dehghani, Josip Djolonga, Basil Mustafa, Piotr Padlewski, Jonathan Heek, Justin Gilmer, Andreas Steiner, Mathilde Caron, Robert Geirhos, Ibrahim Alabdulmohsin, Rodolphe Jenatton, Lucas Beyer, Michael Tschannen, Anurag Arnab, Xiao Wang, Carlos Riquelme, Matthias Minderer, Joan Puigcerver, Utku Evci, Manoj Kumar, Sjoerd van Steenkiste, Gamaleldin~F. Elsayed, Aravindh Mahendran, Fisher Yu, Avital Oliver, Fantine Huot, Jasmijn Bastings, Mark~Patrick Collier, Alexey Gritsenko, Vighnesh Birodkar, Cristina Vasconcelos, Yi~Tay, Thomas Mensink, Alexander Kolesnikov, Filip Pavetić, Dustin Tran, Thomas Kipf, Mario Lučić, Xiaohua Zhai, Daniel Keysers, Jeremiah Harmsen, and Neil Houlsby.
\newblock Scaling vision transformers to 22 billion parameters, 2023.

\bibitem[Esser et~al.(2021)Esser, Rombach, and Ommer]{esser2021taming}
Patrick Esser, Robin Rombach, and Björn Ommer.
\newblock Taming transformers for high-resolution image synthesis, 2021.

\bibitem[Esser et~al.(2024)Esser, Kulal, Blattmann, Entezari, Müller, Saini, Levi, Lorenz, Sauer, Boesel, Podell, Dockhorn, English, Lacey, Goodwin, Marek, and Rombach]{esser2024scaling}
Patrick Esser, Sumith Kulal, Andreas Blattmann, Rahim Entezari, Jonas Müller, Harry Saini, Yam Levi, Dominik Lorenz, Axel Sauer, Frederic Boesel, Dustin Podell, Tim Dockhorn, Zion English, Kyle Lacey, Alex Goodwin, Yannik Marek, and Robin Rombach.
\newblock Scaling rectified flow transformers for high-resolution image synthesis, 2024.

\bibitem[Gafni et~al.(2022)Gafni, Polyak, Ashual, Sheynin, Parikh, and Taigman]{gafni2022makeascene}
Oran Gafni, Adam Polyak, Oron Ashual, Shelly Sheynin, Devi Parikh, and Yaniv Taigman.
\newblock Make-a-scene: Scene-based text-to-image generation with human priors, 2022.

\bibitem[Hessel et~al.(2022)Hessel, Holtzman, Forbes, Bras, and Choi]{hessel2022clipscore}
Jack Hessel, Ari Holtzman, Maxwell Forbes, Ronan~Le Bras, and Yejin Choi.
\newblock Clipscore: A reference-free evaluation metric for image captioning, 2022.

\bibitem[Ho and Salimans(2022)]{ho2022classifierfree}
Jonathan Ho and Tim Salimans.
\newblock Classifier-free diffusion guidance, 2022.

\bibitem[Ho et~al.(2020)Ho, Jain, and Abbeel]{ho2020denoising}
Jonathan Ho, Ajay Jain, and Pieter Abbeel.
\newblock Denoising diffusion probabilistic models.
\newblock \emph{Advances in neural information processing systems}, 33:\penalty0 6840--6851, 2020.

\bibitem[Hoffmann et~al.(2022)Hoffmann, Borgeaud, Mensch, Buchatskaya, Cai, Rutherford, de~Las~Casas, Hendricks, Welbl, Clark, Hennigan, Noland, Millican, van~den Driessche, Damoc, Guy, Osindero, Simonyan, Elsen, Rae, Vinyals, and Sifre]{hoffmann2022training}
Jordan Hoffmann, Sebastian Borgeaud, Arthur Mensch, Elena Buchatskaya, Trevor Cai, Eliza Rutherford, Diego de~Las~Casas, Lisa~Anne Hendricks, Johannes Welbl, Aidan Clark, Tom Hennigan, Eric Noland, Katie Millican, George van~den Driessche, Bogdan Damoc, Aurelia Guy, Simon Osindero, Karen Simonyan, Erich Elsen, Jack~W. Rae, Oriol Vinyals, and Laurent Sifre.
\newblock Training compute-optimal large language models, 2022.

\bibitem[Holtzman et~al.(2020)Holtzman, Buys, Du, Forbes, and Choi]{holtzman2020curious}
Ari Holtzman, Jan Buys, Li~Du, Maxwell Forbes, and Yejin Choi.
\newblock The curious case of neural text degeneration, 2020.

\bibitem[Jin et~al.(2024)Jin, Xu, Xu, Chen, Liao, Tan, Huang, Chen, Lei, Liu, Song, Lei, Zhang, Ou, Gai, and Mu]{jin2024unified}
Yang Jin, Kun Xu, Kun Xu, Liwei Chen, Chao Liao, Jianchao Tan, Quzhe Huang, Bin Chen, Chenyi Lei, An~Liu, Chengru Song, Xiaoqiang Lei, Di~Zhang, Wenwu Ou, Kun Gai, and Yadong Mu.
\newblock Unified language-vision pretraining in llm with dynamic discrete visual tokenization, 2024.

\bibitem[Kaplan et~al.(2020)Kaplan, McCandlish, Henighan, Brown, Chess, Child, Gray, Radford, Wu, and Amodei]{kaplan2020scaling}
Jared Kaplan, Sam McCandlish, Tom Henighan, Tom~B. Brown, Benjamin Chess, Rewon Child, Scott Gray, Alec Radford, Jeffrey Wu, and Dario Amodei.
\newblock Scaling laws for neural language models, 2020.

\bibitem[Karras et~al.(2022)Karras, Aittala, Aila, and Laine]{karras2022elucidating}
Tero Karras, Miika Aittala, Timo Aila, and Samuli Laine.
\newblock Elucidating the design space of diffusion-based generative models, 2022.

\bibitem[Karras et~al.(2024)Karras, Aittala, Lehtinen, Hellsten, Aila, and Laine]{karras2024analyzing}
Tero Karras, Miika Aittala, Jaakko Lehtinen, Janne Hellsten, Timo Aila, and Samuli Laine.
\newblock Analyzing and improving the training dynamics of diffusion models, 2024.

\bibitem[Kingma and Welling(2022)]{kingma2022autoencoding}
Diederik~P Kingma and Max Welling.
\newblock Auto-encoding variational bayes, 2022.

\bibitem[Lipman et~al.(2023)Lipman, Chen, Ben-Hamu, Nickel, and Le]{lipman2023flow}
Yaron Lipman, Ricky T.~Q. Chen, Heli Ben-Hamu, Maximilian Nickel, and Matt Le.
\newblock Flow matching for generative modeling, 2023.

\bibitem[Loshchilov and Hutter(2019)]{loshchilov2019decoupled}
Ilya Loshchilov and Frank Hutter.
\newblock Decoupled weight decay regularization, 2019.

\bibitem[Mei et~al.(2024)Mei, Tu, Delbracio, Talebi, Patel, and Milanfar]{mei2024bigger}
Kangfu Mei, Zhengzhong Tu, Mauricio Delbracio, Hossein Talebi, Vishal~M. Patel, and Peyman Milanfar.
\newblock Bigger is not always better: Scaling properties of latent diffusion models, 2024.

\bibitem[Mentzer et~al.(2023)Mentzer, Minnen, Agustsson, and Tschannen]{mentzer2023finite}
Fabian Mentzer, David Minnen, Eirikur Agustsson, and Michael Tschannen.
\newblock Finite scalar quantization: Vq-vae made simple, 2023.

\bibitem[Peebles and Xie(2023)]{peebles2023scalable}
William Peebles and Saining Xie.
\newblock Scalable diffusion models with transformers, 2023.

\bibitem[Perez et~al.(2017)Perez, Strub, de~Vries, Dumoulin, and Courville]{perez2017film}
Ethan Perez, Florian Strub, Harm de~Vries, Vincent Dumoulin, and Aaron Courville.
\newblock Film: Visual reasoning with a general conditioning layer, 2017.

\bibitem[Pernias et~al.(2023)Pernias, Rampas, Richter, Pal, and Aubreville]{pernias2023wuerstchen}
Pablo Pernias, Dominic Rampas, Mats~L. Richter, Christopher~J. Pal, and Marc Aubreville.
\newblock Wuerstchen: An efficient architecture for large-scale text-to-image diffusion models, 2023.

\bibitem[Radford et~al.(2021)Radford, Kim, Hallacy, Ramesh, Goh, Agarwal, Sastry, Askell, Mishkin, Clark, Krueger, and Sutskever]{radford2021learning}
Alec Radford, Jong~Wook Kim, Chris Hallacy, Aditya Ramesh, Gabriel Goh, Sandhini Agarwal, Girish Sastry, Amanda Askell, Pamela Mishkin, Jack Clark, Gretchen Krueger, and Ilya Sutskever.
\newblock Learning transferable visual models from natural language supervision, 2021.

\bibitem[Rae et~al.(2022)Rae, Borgeaud, Cai, Millican, Hoffmann, Song, Aslanides, Henderson, Ring, Young, Rutherford, Hennigan, Menick, Cassirer, Powell, van~den Driessche, Hendricks, Rauh, Huang, Glaese, Welbl, Dathathri, Huang, Uesato, Mellor, Higgins, Creswell, McAleese, Wu, Elsen, Jayakumar, Buchatskaya, Budden, Sutherland, Simonyan, Paganini, Sifre, Martens, Li, Kuncoro, Nematzadeh, Gribovskaya, Donato, Lazaridou, Mensch, Lespiau, Tsimpoukelli, Grigorev, Fritz, Sottiaux, Pajarskas, Pohlen, Gong, Toyama, de~Masson~d'Autume, Li, Terzi, Mikulik, Babuschkin, Clark, de~Las~Casas, Guy, Jones, Bradbury, Johnson, Hechtman, Weidinger, Gabriel, Isaac, Lockhart, Osindero, Rimell, Dyer, Vinyals, Ayoub, Stanway, Bennett, Hassabis, Kavukcuoglu, and Irving]{rae2022scaling}
Jack~W. Rae, Sebastian Borgeaud, Trevor Cai, Katie Millican, Jordan Hoffmann, Francis Song, John Aslanides, Sarah Henderson, Roman Ring, Susannah Young, Eliza Rutherford, Tom Hennigan, Jacob Menick, Albin Cassirer, Richard Powell, George van~den Driessche, Lisa~Anne Hendricks, Maribeth Rauh, Po-Sen Huang, Amelia Glaese, Johannes Welbl, Sumanth Dathathri, Saffron Huang, Jonathan Uesato, John Mellor, Irina Higgins, Antonia Creswell, Nat McAleese, Amy Wu, Erich Elsen, Siddhant Jayakumar, Elena Buchatskaya, David Budden, Esme Sutherland, Karen Simonyan, Michela Paganini, Laurent Sifre, Lena Martens, Xiang~Lorraine Li, Adhiguna Kuncoro, Aida Nematzadeh, Elena Gribovskaya, Domenic Donato, Angeliki Lazaridou, Arthur Mensch, Jean-Baptiste Lespiau, Maria Tsimpoukelli, Nikolai Grigorev, Doug Fritz, Thibault Sottiaux, Mantas Pajarskas, Toby Pohlen, Zhitao Gong, Daniel Toyama, Cyprien de~Masson~d'Autume, Yujia Li, Tayfun Terzi, Vladimir Mikulik, Igor Babuschkin, Aidan Clark, Diego de~Las~Casas, Aurelia Guy, Chris Jones,
  James Bradbury, Matthew Johnson, Blake Hechtman, Laura Weidinger, Iason Gabriel, William Isaac, Ed~Lockhart, Simon Osindero, Laura Rimell, Chris Dyer, Oriol Vinyals, Kareem Ayoub, Jeff Stanway, Lorrayne Bennett, Demis Hassabis, Koray Kavukcuoglu, and Geoffrey Irving.
\newblock Scaling language models: Methods, analysis and insights from training gopher, 2022.

\bibitem[Rombach et~al.(2022)Rombach, Blattmann, Lorenz, Esser, and Ommer]{rombach2022highresolution}
Robin Rombach, Andreas Blattmann, Dominik Lorenz, Patrick Esser, and Björn Ommer.
\newblock High-resolution image synthesis with latent diffusion models, 2022.

\bibitem[Sauer et~al.(2021)Sauer, Chitta, Müller, and Geiger]{sauer2021projected}
Axel Sauer, Kashyap Chitta, Jens Müller, and Andreas Geiger.
\newblock Projected gans converge faster, 2021.

\bibitem[Sohl-Dickstein et~al.(2015)Sohl-Dickstein, Weiss, Maheswaranathan, and Ganguli]{sohl2015deep}
Jascha Sohl-Dickstein, Eric Weiss, Niru Maheswaranathan, and Surya Ganguli.
\newblock Deep unsupervised learning using nonequilibrium thermodynamics.
\newblock In \emph{International conference on machine learning}, pages 2256--2265. PMLR, 2015.

\bibitem[Song and Ermon(2019)]{song2019generative}
Yang Song and Stefano Ermon.
\newblock Generative modeling by estimating gradients of the data distribution.
\newblock \emph{Advances in neural information processing systems}, 32, 2019.

\bibitem[Song et~al.(2020)Song, Sohl-Dickstein, Kingma, Kumar, Ermon, and Poole]{song2020score}
Yang Song, Jascha Sohl-Dickstein, Diederik~P Kingma, Abhishek Kumar, Stefano Ermon, and Ben Poole.
\newblock Score-based generative modeling through stochastic differential equations.
\newblock \emph{arXiv preprint arXiv:2011.13456}, 2020.

\bibitem[Su et~al.(2023)Su, Lu, Pan, Murtadha, Wen, and Liu]{su2023roformer}
Jianlin Su, Yu~Lu, Shengfeng Pan, Ahmed Murtadha, Bo~Wen, and Yunfeng Liu.
\newblock Roformer: Enhanced transformer with rotary position embedding, 2023.

\bibitem[Tian et~al.(2024)Tian, Jiang, Yuan, Peng, and Wang]{tian2024visual}
Keyu Tian, Yi~Jiang, Zehuan Yuan, Bingyue Peng, and Liwei Wang.
\newblock Visual autoregressive modeling: Scalable image generation via next-scale prediction, 2024.

\bibitem[Touvron et~al.(2023)Touvron, Lavril, Izacard, Martinet, Lachaux, Lacroix, Rozière, Goyal, Hambro, Azhar, Rodriguez, Joulin, Grave, and Lample]{touvron2023llama}
Hugo Touvron, Thibaut Lavril, Gautier Izacard, Xavier Martinet, Marie-Anne Lachaux, Timothée Lacroix, Baptiste Rozière, Naman Goyal, Eric Hambro, Faisal Azhar, Aurelien Rodriguez, Armand Joulin, Edouard Grave, and Guillaume Lample.
\newblock Llama: Open and efficient foundation language models, 2023.

\bibitem[van~den Oord et~al.(2018)van~den Oord, Vinyals, and Kavukcuoglu]{oord2018neural}
Aaron van~den Oord, Oriol Vinyals, and Koray Kavukcuoglu.
\newblock Neural discrete representation learning, 2018.

\bibitem[Vaswani et~al.(2023)Vaswani, Shazeer, Parmar, Uszkoreit, Jones, Gomez, Kaiser, and Polosukhin]{vaswani2023attention}
Ashish Vaswani, Noam Shazeer, Niki Parmar, Jakob Uszkoreit, Llion Jones, Aidan~N. Gomez, Lukasz Kaiser, and Illia Polosukhin.
\newblock Attention is all you need, 2023.

\bibitem[Villegas et~al.(2022)Villegas, Babaeizadeh, Kindermans, Moraldo, Zhang, Saffar, Castro, Kunze, and Erhan]{villegas2022phenaki}
Ruben Villegas, Mohammad Babaeizadeh, Pieter-Jan Kindermans, Hernan Moraldo, Han Zhang, Mohammad~Taghi Saffar, Santiago Castro, Julius Kunze, and Dumitru Erhan.
\newblock Phenaki: Variable length video generation from open domain textual description, 2022.

\bibitem[Wortsman et~al.(2023)Wortsman, Liu, Xiao, Everett, Alemi, Adlam, Co-Reyes, Gur, Kumar, Novak, Pennington, Sohl-dickstein, Xu, Lee, Gilmer, and Kornblith]{wortsman2023smallscale}
Mitchell Wortsman, Peter~J. Liu, Lechao Xiao, Katie Everett, Alex Alemi, Ben Adlam, John~D. Co-Reyes, Izzeddin Gur, Abhishek Kumar, Roman Novak, Jeffrey Pennington, Jascha Sohl-dickstein, Kelvin Xu, Jaehoon Lee, Justin Gilmer, and Simon Kornblith.
\newblock Small-scale proxies for large-scale transformer training instabilities, 2023.

\bibitem[Yu et~al.(2022{\natexlab{a}})Yu, Li, Koh, Zhang, Pang, Qin, Ku, Xu, Baldridge, and Wu]{yu2022vectorquantized}
Jiahui Yu, Xin Li, Jing~Yu Koh, Han Zhang, Ruoming Pang, James Qin, Alexander Ku, Yuanzhong Xu, Jason Baldridge, and Yonghui Wu.
\newblock Vector-quantized image modeling with improved vqgan, 2022{\natexlab{a}}.

\bibitem[Yu et~al.(2022{\natexlab{b}})Yu, Xu, Koh, Luong, Baid, Wang, Vasudevan, Ku, Yang, Ayan, Hutchinson, Han, Parekh, Li, Zhang, Baldridge, and Wu]{yu2022scaling}
Jiahui Yu, Yuanzhong Xu, Jing~Yu Koh, Thang Luong, Gunjan Baid, Zirui Wang, Vijay Vasudevan, Alexander Ku, Yinfei Yang, Burcu~Karagol Ayan, Ben Hutchinson, Wei Han, Zarana Parekh, Xin Li, Han Zhang, Jason Baldridge, and Yonghui Wu.
\newblock Scaling autoregressive models for content-rich text-to-image generation, 2022{\natexlab{b}}.

\bibitem[Yu et~al.(2024)Yu, Lezama, Gundavarapu, Versari, Sohn, Minnen, Cheng, Birodkar, Gupta, Gu, Hauptmann, Gong, Yang, Essa, Ross, and Jiang]{yu2024language}
Lijun Yu, José Lezama, Nitesh~B. Gundavarapu, Luca Versari, Kihyuk Sohn, David Minnen, Yong Cheng, Vighnesh Birodkar, Agrim Gupta, Xiuye Gu, Alexander~G. Hauptmann, Boqing Gong, Ming-Hsuan Yang, Irfan Essa, David~A. Ross, and Lu~Jiang.
\newblock Language model beats diffusion -- tokenizer is key to visual generation, 2024.

\bibitem[Zhai et~al.(2022)Zhai, Kolesnikov, Houlsby, and Beyer]{zhai2022scaling}
Xiaohua Zhai, Alexander Kolesnikov, Neil Houlsby, and Lucas Beyer.
\newblock Scaling vision transformers, 2022.

\end{thebibliography}

\newpage
\appendix
\onecolumn

\pagebreak

\begin{center}
\textbf{\large Supplementary}
\end{center}

\section{Discrete regularizers}
\label{appendix:discrete_regularizers}

To select a simple but performant vector quantization method for our discrete latent space, we compare the classic vector quantization (VQ) \cite{oord2018neural} without additional complexity like codebook reinitialization, lookup free quantization (LFQ) \cite{yu2024language}, and finite scalar quantization (FSQ) \cite{mentzer2023finite}. While training these autoencoders we observe interesting differences in training dynamics with multiple crossing points between FSQ and LFQ for certain metrics. We present those in Figure \ref{fig:quantizers} where we can see that FSQ often takes the lead in the beginning phases of training but eventually gives it up to LFQ. We can also see that both of these methods outperform classic VQ which struggles without additional aids.

\quantizers

\end{document}